%% file: main.tex
\documentclass{article}

\usepackage{arxiv}
\usepackage[utf8]{inputenc} 
\usepackage[T1]{fontenc}    
\usepackage{hyperref}       
\usepackage{url}            
\usepackage{booktabs}       
\usepackage{amsfonts}       
\usepackage{nicefrac}       
\usepackage{microtype}      
\usepackage{lipsum}
\usepackage{graphicx}
\graphicspath{ {./images/} }
\input{math_commands.tex}

\usepackage{natbib}
\usepackage{graphicx}
\usepackage{booktabs}
\usepackage{multirow}
\usepackage{amsmath,amssymb,amsfonts}
\usepackage{amsthm}
\usepackage{mathrsfs}
\usepackage[title]{appendix}
\usepackage{xcolor}
\usepackage{textcomp}
\usepackage{manyfoot}
\usepackage{algorithm}
\usepackage{algorithmicx}
\usepackage{algpseudocode}
\usepackage{listings}
\usepackage{float}
\usepackage{adjustbox,lipsum}
\usepackage{silence}
\ErrorsOff*

\title{Curvature-Aware PCA with Geodesic Tangent Space Aggregation for Semi-Supervised Learning}

\author{
 Alexandre Luis Magalh\~aes Levada\\
  Computing Department\\
  Federal University of S\~ao Carlos\\
  13565-905, S\~ao Carlos-SP, Brazil\\
  \texttt{alexandre.levada@ufscar.br} \\
}

\begin{document}
\maketitle
\begin{abstract}
Principal Component Analysis (PCA) is a fundamental tool for representation learning, but its global linear formulation fails to capture the structure of data supported on curved manifolds. In contrast, manifold learning methods model nonlinearity but often sacrifice the spectral structure and stability of PCA. We propose \emph{Geodesic Tangent Space Aggregation PCA (GTSA-PCA)}, a geometric extension of PCA that integrates curvature awareness and geodesic consistency within a unified spectral framework. Our approach replaces the global covariance operator with curvature-weighted local covariance operators defined over a $k$-nearest neighbor graph, yielding local tangent subspaces that adapt to the manifold while suppressing high-curvature distortions. We then introduce a geodesic alignment operator that combines intrinsic graph distances with subspace affinities to globally synchronize these local representations. The resulting operator admits a spectral decomposition whose leading components define a geometry-aware embedding. We further incorporate semi-supervised information to guide the alignment, improving discriminative structure with minimal supervision. Experiments on real datasets show consistent improvements over PCA, Kernel PCA, Supervised PCA and strong graph-based baselines such as UMAP, particularly in small sample size and high-curvature regimes. Our results position GTSA-PCA as a principled bridge between statistical and geometric approaches to dimensionality reduction.
\end{abstract}

\section{Introduction}\label{sec1}
Dimensionality reduction is a central problem in machine learning and data analysis, aiming to extract compact and informative representations from high-dimensional data while preserving its underlying structure. Classical linear techniques, most notably Principal Component Analysis (PCA) \cite{jolliffe2002principal}, achieve this by identifying directions of maximal variance, offering computational efficiency and strong statistical interpretability. However, such global linear models are fundamentally limited when data lie on or near nonlinear manifolds embedded in high-dimensional spaces. To address this, a rich class of nonlinear methods, often referred to as manifold learning, has been developed, including ISOMAP \cite{tenenbaum2000global}, Locally Linear Embedding (LLE) \cite{roweis2000nonlinear}, Laplacian Eigenmaps \cite{belkin2003laplacian}, DIffusion Maps \cite{COIFMAN2006}, t-SNE \cite{vanDerMaaten2008} and UMAP \cite{mcinnes2018uniform,McInnes2018}. These approaches leverage neighborhood graphs to capture intrinsic geometry, typically relying on geodesic distances or local linear reconstructions. Despite their success, they often suffer from sensitivity to sampling density, graph construction, and noise, and may lack the stability, scalability, or out-of-sample extension properties that make PCA attractive. More recent graph-based and spectral methods attempt to bridge this gap, yet a principled framework that simultaneously preserves local linear structure, respects global manifold geometry, and retains the statistical grounding of PCA remains an open challenge.

In recent years, geometric machine learning has emerged as a powerful paradigm for understanding and exploiting the non-Euclidean structure of data, drawing tools from differential geometry, topology, and spectral theory to design more expressive learning algorithms \cite{bronstein2017geometric}. Within this perspective, data are no longer treated as points in a flat ambient space, but as samples from structured geometric objects such as Riemannian manifolds or graphs endowed with intrinsic metrics. A key insight in this line of research is that curvature encodes fundamental information about local and global geometry, influencing distances, neighborhood relationships, and the behavior of diffusion processes. Consequently, incorporating curvature into learning algorithms has the potential to significantly enhance both dimensionality reduction and metric learning. Recent works have explored curvature-aware constructions in discrete settings, including Ricci curvature on graphs \cite{ollivier2009ricci} and its applications to network analysis and learning \cite{ni2019community}, as well as connections between curvature and robustness or generalization in deep architectures \cite{cao2020curvature}. These developments suggest that curvature is not merely a geometric descriptor, but a functional ingredient that can guide the design of adaptive representations. However, despite these advances, the integration of curvature into classical dimensionality reduction frameworks, particularly in a way that preserves their statistical interpretability and spectral structure, remains largely underexplored.

The central problem addressed in this work is the development of a dimensionality reduction framework that simultaneously preserves the statistical optimality of linear methods and the geometric fidelity required to model data supported on nonlinear manifolds. Classical approaches such as PCA provide efficient and well-understood solutions, but their reliance on a single global linear subspace leads to significant distortions when the underlying data exhibit curvature or complex topology. On the other hand, nonlinear manifold learning techniques capture intrinsic geometry more effectively, yet often lack stability, scalability, and a coherent statistical interpretation. This dichotomy highlights a fundamental gap: existing methods either prioritize global statistical structure or local geometric consistency, but rarely both in a unified and principled manner. Addressing this challenge is particularly important in modern machine learning, where high-dimensional data frequently reside on structured, non-Euclidean domains, and where the quality of learned representations directly impacts downstream tasks such as clustering, classification, and semi-supervised learning. The goal of this work is therefore to design a method that reconciles these competing objectives, enabling geometry-aware, robust, and discriminatively meaningful representations.

Building on these observations, we propose Geodesic Tangent Space Aggregation PCA (GTSA-PCA), a novel dimensionality reduction framework that overcomes key limitations of both classical PCA and existing manifold learning methods. Unlike PCA, which relies on a single global covariance operator, and unlike manifold methods that often lack a coherent statistical interpretation, GTSA-PCA introduces a two-stage geometric construction that unifies local linear modeling with global nonlinear structure. Specifically, the method replaces the global covariance matrix with a collection of curvature-aware local covariance operators defined over a k-nearest neighbor graph, yielding reliable tangent space approximations even in highly curved regions. These local models are then globally synchronized through a geodesic alignment operator that couples intrinsic graph distances with subspace affinities, leading to a well-defined spectral problem whose solution provides a consistent low-dimensional embedding. This formulation is fundamentally different from existing approaches such as Locally Linear Embedding or Laplacian-based methods, as it explicitly models the interaction between curvature, geodesic structure, and subspace alignment within a single operator-theoretic framework. The main contributions of this work are threefold: (i) the introduction of curvature-weighted local covariance estimation for robust tangent space recovery; (ii) a novel geodesic alignment operator that enables the global aggregation of local linear models; and (iii) a semi-supervised extension that incorporates label information to guide the embedding toward discriminative structures. Together, these elements establish GTSA-PCA as a principled and flexible generalization of PCA to nonlinear, graph-structured domains.

The remainder of this paper is organized as follows. In Section 2, Theoretical Background, we review the fundamental concepts underlying our approach, including classical PCA, manifold learning, graph-based representations, and notions of curvature in differential geometry. Section 3, Geodesic Tangent Space Aggregation PCA, presents the proposed method in detail, describing both the curvature-aware local modeling and the global geodesic alignment framework, along with its semi-supervised extension. In Section 4, Computational Experiments and Results, we evaluate the performance of GTSA-PCA on a variety of synthetic and real-world datasets, comparing it against established dimensionality reduction and graph-based techniques. Finally, Section 5, Conclusions and Final Remarks, summarizes the main findings, discusses limitations, and outlines directions for future research.

\section{Related Work}

Dimensionality reduction remains a central problem in machine learning, with classical methods such as Principal Component Analysis (PCA) providing a statistically optimal linear framework. Extensions such as Kernel PCA \cite{Scholkopf1998} introduce nonlinearity via implicit feature mappings, but still rely on global operators and predefined kernels, limiting their ability to capture complex geometric structures. This limitation has motivated a large body of work on manifold learning, where the goal is to recover low-dimensional representations that respect the intrinsic geometry of the data.

Early manifold learning methods, including Laplacian Eigenmaps \cite{belkin2003laplacian}, Diffusion Maps \cite{COIFMAN2006}, Local Tangent Space Alignment (LTSA) \cite{zhang2004principal}, and Hessian Eigenmaps \cite{Donoho2003}, introduced the idea of leveraging local neighborhoods and graph-based constructions to approximate nonlinear manifolds. While these approaches successfully capture local geometry, they often rely on pairwise distances or local linear reconstructions and do not explicitly model higher-order geometric properties such as curvature. Moreover, their connection to classical statistical methods such as PCA remains indirect, as they lack a unified covariance-based interpretation.

More recent work has focused on graph-based and spectral approaches that aim to learn representations by exploiting the structure of data graphs. For instance, graph learning frameworks have been proposed to jointly infer graph topology and low-dimensional embeddings, improving robustness to noise and sampling density variations \cite{Dong2019, Xia2021}. In parallel, advances in geometric deep learning have emphasized the importance of non-Euclidean structure, leading to architectures that operate on graphs and manifolds \cite{bronstein2017geometric}. These methods have demonstrated strong empirical performance, but typically involve complex models and do not directly generalize classical linear techniques such as PCA.

A particularly relevant line of research explores the incorporation of curvature into learning algorithms. Recent studies have shown that curvature-based quantities, such as Ricci curvature, can provide valuable insights into graph structure and improve representation learning, especially in the context of graph neural networks \cite{Topping2022, Baptista2024}. Similarly, approaches based on Riemannian geometry and non-Euclidean embeddings, including hyperbolic and metric learning frameworks, have highlighted the importance of intrinsic geometry for capturing hierarchical and nonlinear relationships \cite{Chami2019, Papillon2025}. Despite these advances, most existing methods either incorporate curvature implicitly or rely on complex optimization schemes, without providing a direct integration into a PCA-like spectral framework.

Closer to our approach, recent works have revisited local geometric modeling by incorporating curvature-aware mechanisms into manifold learning. For example, curvature-regularized embeddings and adaptive local geometry methods have been proposed to improve the stability and expressiveness of local models \cite{Li2018, Zou2026}. However, these methods typically focus on local reconstruction or regularization and do not address the global alignment of local tangent spaces in a principled manner. In particular, they lack a unifying operator that combines curvature, geodesic structure, and subspace alignment within a single spectral formulation.

In contrast, the proposed GTSA-PCA provides a unified framework that bridges these gaps. By combining curvature-aware local covariance estimation with a geodesic alignment operator defined over a data graph, our method extends PCA to nonlinear domains while preserving its spectral and statistical foundations. This positions GTSA-PCA at the intersection of classical dimensionality reduction, manifold learning, and geometric machine learning, offering a principled approach to integrating first- and second-order geometric information into representation learning.

\section{Theoretical background}

In this section, we introduce the theoretical foundations necessary for a comprehensive understanding of the proposed Geodesic Tangent Space Aggregation PCA (GTSA-PCA). We begin by revisiting classical Principal Component Analysis and its interpretation in terms of covariance operators and spectral decomposition, establishing the baseline from which our formulation departs. We then discuss key concepts from differential geometry such as fundamental forms, shape operator and mean curvature. Together, these elements form the conceptual and mathematical basis for the proposed framework.

\subsection{Principal Component Analysis}

Principal Component Analysis (PCA) is a foundational technique in statistical learning and signal processing, providing a principled approach to linear dimensionality reduction through the spectral analysis of second-order statistics. Formally, PCA can be interpreted as the empirical realization of the Karhunen--Lo\`eve transform, whereby a random vector $\mathbf{x} \in \mathbb{R}^m$ is expanded in the orthonormal eigenbasis of its covariance operator. Given a dataset, PCA estimates this operator directly from samples and identifies the principal directions as those maximizing the projected variance. This leads to a low-dimensional representation obtained via orthogonal projection onto the leading eigenvectors \cite{jolliffe2002principal}.

Notably, PCA admits two equivalent optimality characterizations: it yields the rank-$d$ linear subspace that maximizes retained variance, and simultaneously minimizes the mean squared reconstruction error among all linear projections of the same dimension \cite{cunningham15a}. From a statistical perspective, PCA performs decorrelation by diagonalizing the covariance matrix, effectively transforming the data into a set of uncorrelated components ordered by their explained variance. Its nonparametric nature, requiring no explicit assumptions on the underlying data distribution beyond finite second moments, combined with its closed-form solution via eigenvalue decomposition, has made PCA a ubiquitous tool for compression, denoising, and feature extraction. However, its reliance on global linear structure inherently limits its ability to capture nonlinear geometric patterns, motivating the development of manifold-aware generalizations \cite{Jolliffe2016}.

Let $\mathbf{Z} = [\mathbf{T} \ \mathbf{S}] \in \mathbb{R}^{m \times m}$ be an orthonormal basis for $\mathbb{R}^m$, such that $\mathbf{Z}^\top \mathbf{Z} = \mathbf{I}$. The matrix $\mathbf{T} = [\mathbf{w}_1, \mathbf{w}_2, \ldots, \mathbf{w}_d] \in \mathbb{R}^{m \times d}$ spans the $d$-dimensional principal subspace to be retained, while $\mathbf{S} = [\mathbf{w}_{d+1}, \mathbf{w}_{d+2}, \ldots, \mathbf{w}_m] \in \mathbb{R}^{m \times (m-d)}$ spans its orthogonal complement, corresponding to the subspace discarded during dimensionality reduction. In this formulation, PCA seeks an optimal orthogonal decomposition of the ambient space into informative and non-informative components with respect to variance.

Let $\mathbf{Z} = [\mathbf{T} \ \mathbf{S}] \in \mathbb{R}^{m \times m}$ be an orthonormal basis for $\mathbb{R}^m$, satisfying $\mathbf{Z}^\top \mathbf{Z} = \mathbf{I}$. The matrix $\mathbf{T} = [\mathbf{w}_1, \mathbf{w}_2, \ldots, \mathbf{w}_d] \in \mathbb{R}^{m \times d}$ spans the $d$-dimensional principal subspace to be retained, with $d < m$, while $\mathbf{S} = [\mathbf{w}_{d+1}, \ldots, \mathbf{w}_m] \in \mathbb{R}^{m \times (m-d)}$ spans its orthogonal complement. In this decomposition, $\mathbf{T}$ captures the informative directions of maximal variance, whereas $\mathbf{S}$ corresponds to the subspace discarded during dimensionality reduction.

Given a centered dataset $\mathbf{X} = [\mathbf{x}_1, \mathbf{x}_2, \ldots, \mathbf{x}_n] \in \mathbb{R}^{m \times n}$, i.e., $\frac{1}{n}\sum_{i=1}^n \mathbf{x}_i = \mathbf{0}$, the goal of PCA is to determine an orthonormal set of directions $\{\mathbf{w}_j\}_{j=1}^d$ that maximizes the variance of the projected data. Any vector $\mathbf{x} \in \mathbb{R}^m$ can be expanded in the basis $\mathbf{Z}$ as
\begin{equation}
	\mathbf{x} = \sum_{j=1}^{m} (\mathbf{w}_j^\top \mathbf{x}) \mathbf{w}_j = \sum_{j=1}^{m} c_j \mathbf{w}_j,
\end{equation}
where $c_j = \mathbf{w}_j^\top \mathbf{x}$ are the projection coefficients. The low-dimensional representation $\mathbf{y} \in \mathbb{R}^d$ is obtained via the linear transformation
\begin{equation}
	\mathbf{y} = \mathbf{T}^\top \mathbf{x},
\end{equation}
which, due to orthonormality ($\mathbf{w}_i^\top \mathbf{w}_j = \delta_{ij}$), yields
\begin{equation}
	\mathbf{y} = [c_1, c_2, \ldots, c_d]^\top.
\end{equation}

PCA seeks the transformation $\mathbf{T}$ that maximizes the total variance of the projected data, leading to the objective functional
\begin{equation}
	J_{\text{PCA}}(\mathbf{T}) = \mathbb{E}\left[\|\mathbf{y}\|^2\right] = \sum_{j=1}^{d} \mathbb{E}[c_j^2].
\end{equation}
Since $c_j = \mathbf{w}_j^\top \mathbf{x}$, this can be rewritten as
\begin{equation}
	J_{\text{PCA}}(\mathbf{T}) = \sum_{j=1}^{d} \mathbf{w}_j^\top \boldsymbol{\Sigma} \mathbf{w}_j,
\end{equation}
where $\boldsymbol{\Sigma} = \mathbb{E}[\mathbf{x}\mathbf{x}^\top]$ denotes the covariance matrix of the data. The resulting optimization problem is
\begin{equation}
	\max_{\{\mathbf{w}_j\}_{j=1}^d} \sum_{j=1}^{d} \mathbf{w}_j^\top \boldsymbol{\Sigma} \mathbf{w}_j
	\quad \text{subject to} \quad \|\mathbf{w}_j\| = 1, \ \mathbf{w}_i^\top \mathbf{w}_j = 0 \ (i \neq j).
\end{equation}

Introducing Lagrange multipliers, one obtains the stationarity condition
\begin{equation}
	\boldsymbol{\Sigma} \mathbf{w}_j = \lambda_j \mathbf{w}_j,
\end{equation}
which corresponds to the eigenvalue problem associated with the covariance matrix. Therefore, the optimal solution is given by selecting the $d$ eigenvectors corresponding to the largest eigenvalues $\{\lambda_j\}_{j=1}^d$. This ensures maximal variance preservation and, equivalently, minimal reconstruction error among all rank-$d$ linear approximations.


PCA admits a natural geometric interpretation as an orthogonal transformation that reorients the coordinate system to align with the directions of maximal variance in the data. Given a dataset $\mathbf{X} \in \mathbb{R}^{m \times n}$, the first step consists of centering the data by subtracting the empirical mean, ensuring that the distribution is aligned around the origin. The transformation then projects each sample $\mathbf{x} \in \mathbb{R}^m$ onto the eigenbasis of the covariance matrix $\boldsymbol{\Sigma}$, effectively eliminating second-order statistical dependencies (i.e., correlations) between the original features.

Geometrically, this procedure corresponds to a rotation of the coordinate axes such that the new axes, defined by the eigenvectors of $\boldsymbol{\Sigma}$, are aligned with the principal directions of data dispersion. In this rotated coordinate system, the covariance matrix becomes diagonal, and the variance along each axis is given by the corresponding eigenvalue. The ordering of these eigenvalues induces a natural hierarchy of directions, where the leading components capture the most significant modes of variation. Dimensionality reduction is then achieved by truncating this basis, retaining only the directions associated with the largest eigenvalues. This truncation yields the optimal low-dimensional linear approximation in the mean squared error sense.

Under the assumption that the data are drawn from a multivariate Gaussian distribution, PCA admits an additional probabilistic interpretation. In this case, the data distribution can be visualized as a hyperellipsoid whose principal axes coincide with the eigenvectors of $\boldsymbol{\Sigma}$, and whose axis lengths are proportional to the square roots of the eigenvalues. The principal components correspond to projections onto these axes, and directions associated with small eigenvalues contribute negligibly to the overall data variability, thus justifying their removal. Even beyond the Gaussian setting, this geometric intuition remains valid and provides insight into why PCA effectively captures the dominant structure of high-dimensional data.


Despite its widespread use and strong theoretical foundations, Principal Component Analysis (PCA) exhibits several well-known limitations that restrict its applicability in modern machine learning settings, particularly when dealing with complex, high-dimensional, and nonlinear data.

A fundamental limitation of PCA lies in its \emph{global linearity}. The method assumes that the data can be well-approximated by a single low-dimensional linear subspace embedded in the ambient space. While this assumption is reasonable for data with approximately linear structure, it becomes inadequate when the data lie on or near a nonlinear manifold. In such cases, PCA fails to capture intrinsic geometric properties, often leading to distortions that obscure meaningful patterns. This limitation has motivated the development of manifold learning techniques that explicitly account for nonlinear structure.

Another important drawback is that PCA relies exclusively on \emph{second-order statistics}, as it is entirely determined by the covariance matrix. Consequently, it is insensitive to higher-order dependencies and may fail to capture non-Gaussian structures present in the data. In scenarios where relevant information is encoded in higher-order moments or complex feature interactions, PCA may provide suboptimal representations.

PCA is also known to be \emph{sensitive to scaling and outliers}. Since it maximizes variance, features with larger scales tend to dominate the principal components unless appropriate normalization is applied. Moreover, the covariance matrix is highly sensitive to extreme values, meaning that outliers can significantly distort the estimated principal directions, degrading the quality of the embedding.

From a geometric standpoint, PCA does not incorporate any notion of \emph{locality} or \emph{intrinsic geometry}. The method treats all data points uniformly, ignoring neighborhood relationships that are often crucial for capturing the structure of data supported on manifolds. As a result, PCA cannot preserve geodesic distances or local neighborhoods, which are essential in many applications such as visualization, clustering, and semi-supervised learning.

Finally, PCA is inherently \emph{unsupervised} and does not leverage label information when available. While this makes it broadly applicable, it also means that the resulting representation may not be optimal for downstream discriminative tasks. Extensions such as supervised or semi-supervised variants attempt to address this issue, but they often sacrifice the simplicity and interpretability of the original formulation.

These limitations highlight the need for more expressive dimensionality reduction frameworks that retain the favorable properties of PCA, such as its spectral formulation and statistical interpretabilit, while incorporating nonlinear geometry, locality, and, when available, supervisory information.

\subsection{Differential Geometry Basics}

In this subsection, we briefly review fundamental concepts from differential geometry that underpin the proposed framework, providing the mathematical tools necessary to describe and analyze data supported on nonlinear manifolds \cite{spivak1999comprehensive,do_carmo_differential_2016,tu2017differential,needham2021visual}. In contrast to classical Euclidean assumptions, many high-dimensional datasets encountered in practice can be more accurately modeled as samples from a low-dimensional smooth manifold $\mathcal{M} \subset \mathbb{R}^m$. Within this setting, the notion of a \emph{tangent space} $T_{\mathbf{x}}\mathcal{M}$ at a point $\mathbf{x} \in \mathcal{M}$ plays a central role, as it provides a first-order linear approximation of the manifold in a local neighborhood. This approximation is formally characterized by the \emph{first fundamental form}, which induces a Riemannian metric and enables the measurement of lengths, angles, and local distances. However, first-order information alone is insufficient to capture the intrinsic geometry of curved spaces. To this end, the \emph{second fundamental form} and the associated \emph{shape operator} encode how the manifold bends within the ambient space, providing a precise quantification of local curvature. These curvature descriptors govern deviations from linearity and directly influence the reliability of local Euclidean approximations. In the context of dimensionality reduction, such geometric quantities offer a principled mechanism to modulate local models, guiding the construction of representations that adapt to the underlying curvature of the data manifold. These concepts form the geometric backbone of our method, enabling the integration of curvature-aware weighting and tangent space alignment within a unified framework.

\paragraph{Riemannian manifold}
A \emph{Riemannian manifold} is a pair $(\mathcal{M}, g)$, where $\mathcal{M}$ is a smooth $d$-dimensional manifold and $g$ is a Riemannian metric, that is, a smoothly varying inner product defined on the tangent space at each point. Formally, for every $\mathbf{x} \in \mathcal{M}$, the metric $g_{\mathbf{x}}: T_{\mathbf{x}}\mathcal{M} \times T_{\mathbf{x}}\mathcal{M} \rightarrow \mathbb{R}$ is a symmetric, positive-definite bilinear form that varies smoothly with $\mathbf{x}$. This structure allows one to generalize fundamental geometric notions, such as lengths, angles, distances, and volumes, to curved spaces \cite{docarmo1992riemannian, gallot2004riemannian, tu2011manifolds, lee2013smooth, petersen2016riemannian, jost2017riemannian, lee2018riemannian}. In particular, the length of a smooth curve $\gamma: [0,1] \rightarrow \mathcal{M}$ is defined as
\[
L(\gamma) = \int_0^1 \sqrt{g_{\gamma(t)}\big(\dot{\gamma}(t), \dot{\gamma}(t)\big)} \, dt,
\]
and the geodesic distance between two points is given by the length of the shortest curve connecting them.

The notion of a Riemannian manifold is central in geometric machine learning because it provides a mathematically rigorous framework for modeling data with intrinsic nonlinear structure. In many applications, high-dimensional observations are assumed to lie near a low-dimensional manifold embedded in $\mathbb{R}^m$, where Euclidean distances fail to capture the true relationships between data points. By endowing the manifold with a Riemannian metric, one can define intrinsic distances (geodesics), local neighborhoods, and curvature, all of which are essential for tasks such as dimensionality reduction, clustering, and metric learning.

Moreover, the Riemannian framework justifies the use of local linear approximations via tangent spaces, which serve as first-order models of the manifold. It also enables the incorporation of higher-order geometric information, such as curvature, which quantifies deviations from flatness and directly impacts the reliability of these local approximations. Many modern algorithms, such as manifold learning methods, graph-based techniques, and geometric deep learning models, implicitly or explicitly rely on these concepts. In this sense, Riemannian geometry provides the theoretical foundation for designing learning algorithms that are sensitive to the intrinsic structure of data, rather than the ambient Euclidean space in which they are represented.

\paragraph{Tangent Spaces and Local Linear Structure}

\begin{figure}
	\centering
	\includegraphics[scale=0.2]{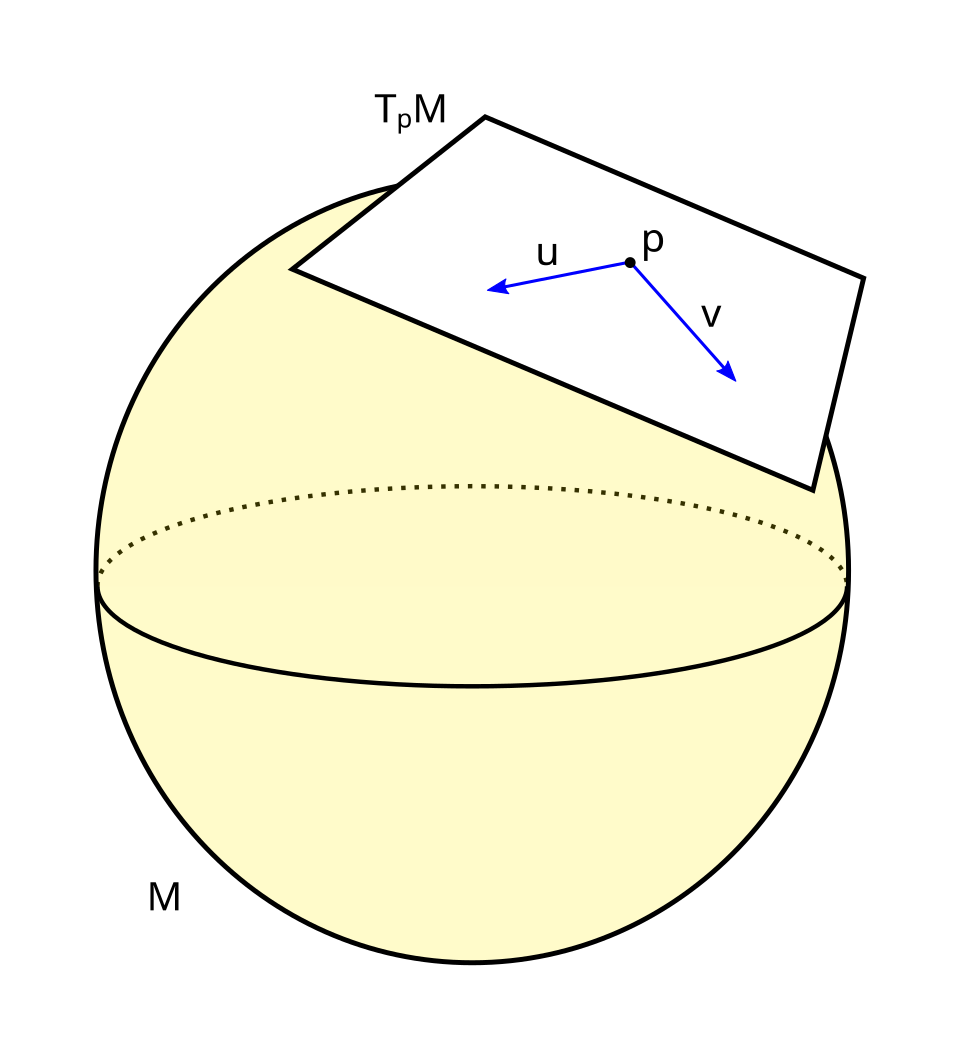}
	\caption{Illustration of the tangent space at a point $\mathbf{x}$ on a sphere embedded in $\mathbb{R}^3$. The tangent plane $T_{\mathbf{x}}\mathcal{M}$ provides a first-order linear approximation of the manifold in a local neighborhood of $\mathbf{x}$, capturing the directions of permissible infinitesimal variations along the surface.}
	\label{fig:tangent}
\end{figure}

Let $\mathcal{M} \subset \mathbb{R}^m$ be a smooth $d$-dimensional manifold and let $\mathbf{x} \in \mathcal{M}$. The \emph{tangent space} at $\mathbf{x}$, denoted by $T_{\mathbf{x}}\mathcal{M}$, is a $d$-dimensional vector space that provides the best linear approximation of the manifold in a neighborhood of $\mathbf{x}$. Formally, $T_{\mathbf{x}}\mathcal{M}$ can be defined as the set of tangent vectors to all smooth curves $\gamma: (-\epsilon,\epsilon) \rightarrow \mathcal{M}$ such that $\gamma(0) = \mathbf{x}$, i.e.,
\[
T_{\mathbf{x}}\mathcal{M} = \left\{ \dot{\gamma}(0) \in \mathbb{R}^m \ \middle| \ \gamma(0) = \mathbf{x}, \ \gamma(t) \in \mathcal{M} \right\}.
\]
If $\mathcal{M}$ is locally parameterized by a smooth map $\boldsymbol{\phi}: \mathbb{R}^d \rightarrow \mathcal{M}$, then the tangent space at $\mathbf{x} = \boldsymbol{\phi}(\mathbf{u})$ is given by the span of the Jacobian vectors:
\[
T_{\mathbf{x}}\mathcal{M} = \mathrm{span} \left\{ \frac{\partial \boldsymbol{\phi}}{\partial u_1}, \frac{\partial \boldsymbol{\phi}}{\partial u_2}, \ldots, \frac{\partial \boldsymbol{\phi}}{\partial u_d} \right\}.
\]
Figure \ref{fig:tangent} shows an illustration of a tangent plane on a sphere. In the context of geometric machine learning, tangent spaces play a fundamental role as they enable the approximation of nonlinear structures by locally linear models. Many algorithms, including manifold learning methods and graph-based approaches, rely on the assumption that, within sufficiently small neighborhoods, the data can be well approximated by a linear subspace. This justifies the use of techniques such as local PCA to estimate $T_{\mathbf{x}}\mathcal{M}$ from sampled data.

From a geometric perspective, tangent spaces capture first-order information about the data manifold, allowing one to define local coordinates, project data points, and approximate geodesic distances. Moreover, they provide the foundation for incorporating higher-order geometric quantities, such as curvature, which describe how the manifold deviates from its tangent approximation. In modern learning frameworks, tangent space estimation is a key component in constructing representations that preserve local structure while enabling global inference, particularly in settings involving dimensionality reduction, semi-supervised learning, and metric learning on non-Euclidean domains.

\paragraph{First Fundamental Form and Riemannian Metric}

Let $\mathcal{M} \subset \mathbb{R}^m$ be a smooth $d$-dimensional manifold and let $\boldsymbol{\phi}: U \subset \mathbb{R}^d \rightarrow \mathcal{M}$ be a local parametrization. The \emph{first fundamental form}, also known as the \emph{Riemannian metric tensor}, is defined as the inner product induced on the tangent space $T_{\mathbf{x}}\mathcal{M}$ by the ambient Euclidean space. In local coordinates, it is given by
\[
g_{ij}(\mathbf{u}) = \left\langle \frac{\partial \boldsymbol{\phi}}{\partial u_i}, \frac{\partial \boldsymbol{\phi}}{\partial u_j} \right\rangle, \quad i,j = 1, \ldots, d,
\]
where $\mathbf{x} = \boldsymbol{\phi}(\mathbf{u}) \in \mathcal{M}$ and $\langle \cdot, \cdot \rangle$ denotes the standard inner product in $\mathbb{R}^m$. The matrix $\mathbf{G}(\mathbf{u}) = [g_{ij}(\mathbf{u})] \in \mathbb{R}^{d \times d}$ is symmetric and positive definite, and it encodes all intrinsic geometric information related to lengths and angles on the manifold.

The first fundamental form enables the computation of geometric quantities such as the length of curves, angles between tangent vectors, and local distances. In particular, for a smooth curve $\gamma(t) = \boldsymbol{\phi}(\mathbf{u}(t))$ on $\mathcal{M}$, its length is given by
\[
L(\gamma) = \int \sqrt{\dot{\mathbf{u}}(t)^\top \mathbf{G}(\mathbf{u}(t)) \dot{\mathbf{u}}(t)} \, dt,
\]
which generalizes the notion of Euclidean arc length to curved spaces.

In geometric machine learning, the first fundamental form plays a central role as it defines the intrinsic metric structure of the data manifold. Unlike ambient Euclidean distances, which may be misleading in high-dimensional settings, the metric tensor captures the true local geometry of the data. This is particularly important for algorithms that rely on neighborhood relationships, distance computations, or diffusion processes, such as manifold learning, graph-based methods, and spectral techniques.

Moreover, the metric tensor provides the foundation for constructing geodesic distances and Laplace--Beltrami operators, both of which are fundamental tools in modern representation learning. In the context of dimensionality reduction, the first fundamental form justifies the use of local linear approximations (via tangent spaces) while also highlighting their limitations in regions of high curvature. Consequently, it serves as a key ingredient for designing geometry-aware methods that adapt to the intrinsic structure of the data, as pursued in the proposed framework.

\paragraph{Second Fundamental Form and Curvature}

Let $\mathcal{M} \subset \mathbb{R}^m$ be a smooth $d$-dimensional manifold embedded in Euclidean space, and let $\mathbf{x} \in \mathcal{M}$. While the first fundamental form captures the intrinsic geometry of $\mathcal{M}$, the \emph{second fundamental form} encodes how the manifold bends within the ambient space. Formally, the second fundamental form at $\mathbf{x}$ is a symmetric bilinear map
\[
\mathrm{II}_{\mathbf{x}} : T_{\mathbf{x}}\mathcal{M} \times T_{\mathbf{x}}\mathcal{M} \rightarrow N_{\mathbf{x}}\mathcal{M},
\]
where $N_{\mathbf{x}}\mathcal{M}$ denotes the normal space at $\mathbf{x}$. Intuitively, $\mathrm{II}_{\mathbf{x}}(\mathbf{u}, \mathbf{v})$ measures the variation of the tangent space along directions $\mathbf{u}, \mathbf{v} \in T_{\mathbf{x}}\mathcal{M}$, projecting this variation onto the normal space.

In local coordinates, let $\boldsymbol{\phi}: U \subset \mathbb{R}^d \rightarrow \mathcal{M}$ be a parametrization and denote by $\mathbf{n}$ a unit normal vector field. The second fundamental form can be expressed as
\[
\mathrm{II}_{ij} = \left\langle \frac{\partial^2 \boldsymbol{\phi}}{\partial u_i \partial u_j}, \mathbf{n} \right\rangle,
\]
which captures second-order variations of the embedding. In geometric machine learning, the second fundamental form plays a crucial role by providing a quantitative measure of local nonlinearity. While tangent spaces offer first-order (linear) approximations of the manifold, curvature captures the deviation from these approximations, indicating how rapidly the manifold bends in different directions. This information is essential for assessing the validity of local linear models and for designing adaptive methods that account for geometric complexity.

In particular, curvature can guide the weighting of neighborhood relationships, the construction of graph-based representations, and the selection of reliable local regions for learning. Regions of high curvature often correspond to areas where linear assumptions break down, suggesting that their influence should be attenuated. Conversely, low-curvature regions are well-approximated by tangent spaces and can be exploited more confidently. Incorporating curvature information into learning algorithms therefore enables a more faithful representation of the underlying data geometry, improving robustness and performance in tasks such as dimensionality reduction, clustering, and semi-supervised learning.

\subsection{Shape Operator and Curvature Structure}

Let $\mathcal{M} \subset \mathbb{R}^m$ be a smooth embedded manifold and let $\mathbf{x} \in \mathcal{M}$. The \emph{shape operator} (or \emph{Weingarten map}) at $\mathbf{x}$ is a linear operator $\mathbf{S}_{\mathbf{x}}: T_{\mathbf{x}}\mathcal{M} \rightarrow T_{\mathbf{x}}\mathcal{M}$ that characterizes how the normal direction to the manifold varies along tangent directions. Formally, given a unit normal vector field $\mathbf{n}$ defined in a neighborhood of $\mathbf{x}$, the shape operator is defined as
\[
\mathbf{S}_{\mathbf{x}}(\mathbf{v}) = - D_{\mathbf{v}} \mathbf{n},
\]
where $D_{\mathbf{v}} \mathbf{n}$ denotes the directional derivative of the normal vector field in the direction $\mathbf{v} \in T_{\mathbf{x}}\mathcal{M}$. Intuitively, $\mathbf{S}_{\mathbf{x}}$ measures how the normal vector ``rotates'' as one moves along the surface, thereby encoding the curvature of the manifold.

The shape operator is self-adjoint with respect to the Riemannian metric and is closely related to the second fundamental form through the relation
\[
\langle \mathbf{S}_{\mathbf{x}} \mathbf{u}, \mathbf{v} \rangle = \langle \mathrm{II}_{\mathbf{x}}(\mathbf{u}, \mathbf{v}), \mathbf{n} \rangle,
\]
for all $\mathbf{u}, \mathbf{v} \in T_{\mathbf{x}}\mathcal{M}$. Its eigenvalues $\{\kappa_i\}_{i=1}^d$ are the \emph{principal curvatures}, and the corresponding eigenvectors define the \emph{principal directions}. These quantities provide a complete second-order characterization of the local geometry, capturing both the magnitude and orientation of curvature.

In the context of geometric machine learning, the shape operator offers a compact and informative representation of local geometric complexity. While tangent spaces encode first-order structure, the shape operator captures second-order variations, enabling a more precise assessment of how well local linear models approximate the underlying manifold. This is particularly relevant in high-curvature regions, where naive linear approximations may lead to significant distortions.

From an algorithmic perspective, the eigenstructure of $\mathbf{S}_{\mathbf{x}}$ can be used to guide adaptive learning strategies. For instance, curvature information derived from the shape operator can inform the weighting of neighborhood relationships, the construction of graph affinities, or the selection of reliable local patches. In dimensionality reduction and metric learning, incorporating such curvature-aware mechanisms leads to representations that better respect the intrinsic geometry of the data, improving both stability and discriminative power. In this sense, the shape operator serves as a key bridge between differential geometry and modern data-driven learning frameworks.

\begin{figure}
	\centering
	\includegraphics[scale=0.6]{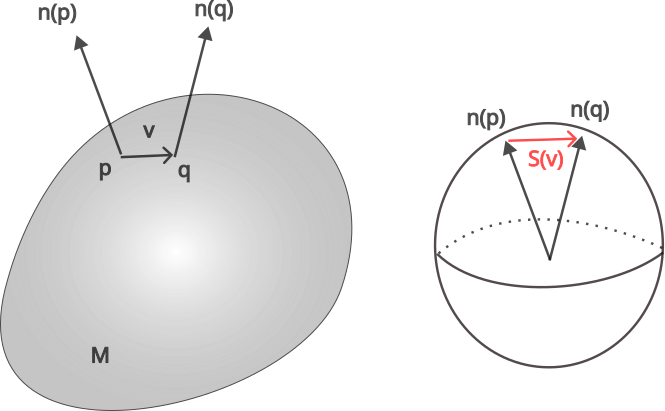}
	\caption{The shape operator measures how a normal vector changes from the tail to the tip of a tangent vector $\vec{v}$. In other words, it tells us how fast the normal changes, or roughly speaking, how fast a surface turns if you roll it along the floor in some direction.}
	\label{fig:shape}
\end{figure}

Figure~\ref{fig:shape} illustrates how the shape operator captures the directional variation of the normal vector across a curved surface, revealing how local curvature depends on the chosen tangent direction.

The concepts introduced in this section establish the geometric foundation upon which the proposed method is built. By modeling data as samples from a smooth manifold, we move beyond purely Euclidean assumptions and adopt a representation that reflects the intrinsic structure of the underlying space. In this context, tangent spaces provide local linear approximations, the first fundamental form defines the intrinsic metric governing distances and angles, and the second fundamental form together with the shape operator captures the curvature and deviation from linearity.

These geometric constructs are not merely theoretical abstractions; they play a direct role in the design of modern learning algorithms. In particular, they enable the development of methods that adapt to local structure, quantify geometric complexity, and reconcile local linearity with global nonlinear behavior. This perspective is especially relevant in high-dimensional settings, where the ambient space can obscure the true relationships between data points.

The proposed Geodesic Tangent Space Aggregation PCA builds directly upon these principles. By leveraging tangent space estimation, curvature-aware modeling, and geodesic relationships on graphs, the method seeks to integrate first- and second-order geometric information into a unified dimensionality reduction framework. The next section formalizes this approach, translating the geometric insights discussed here into a concrete algorithmic construction.

\subsection{Wasserstein Distance}

Comparing probability distributions is a fundamental problem in machine learning, arising in tasks such as generative modeling, domain adaptation, clustering, and representation learning. Classical divergence measures, such as the Kullback--Leibler (KL) divergence or Jensen--Shannon divergence, rely on pointwise comparisons of densities and often fail when the supports of the distributions do not overlap. In contrast, the \emph{Wasserstein distance}, rooted in optimal transport theory, provides a geometrically meaningful notion of distance by explicitly accounting for the cost of transporting mass between distributions \cite{Villani2008,Ambrosio2008}. This makes it particularly well-suited for applications involving structured data, where the geometry of the underlying space plays a crucial role.

Let $(\mathcal{X}, d)$ be a metric space and let $\mu, \nu$ be two probability measures defined on $\mathcal{X}$. The $p$-Wasserstein distance between $\mu$ and $\nu$ is defined as
\begin{equation}
	W_p(\mu, \nu) = \left( \inf_{\gamma \in \Pi(\mu, \nu)} 
	\int_{\mathcal{X} \times \mathcal{X}} d(x, y)^p \, d\gamma(x,y) \right)^{\frac{1}{p}},
\end{equation}
where $\Pi(\mu, \nu)$ denotes the set of all couplings (joint distributions) $\gamma$ with marginals $\mu$ and $\nu$. Intuitively, $\gamma(x,y)$ represents how much mass is transported from $x$ to $y$, and the Wasserstein distance corresponds to the minimum transportation cost required to transform $\mu$ into $\nu$. In the discrete setting, where \cite{Peyre2019}
\[
\mu = \sum_{i=1}^n a_i \delta_{x_i}, \quad
\nu = \sum_{j=1}^m b_j \delta_{y_j},
\]
the Wasserstein distance reduces to the solution of the following optimal transport problem:
\begin{equation}
	\min_{\mathbf{T} \in \mathbb{R}^{n \times m}} 
	\sum_{i=1}^n \sum_{j=1}^m T_{ij} \, d(x_i, y_j)^p,
\end{equation}
subject to
\begin{equation}
	\sum_{j=1}^m T_{ij} = a_i, \quad
	\sum_{i=1}^n T_{ij} = b_j, \quad
	T_{ij} \geq 0,
\end{equation}
where $\mathbf{T}$ is the transport plan.

One of the main advantages of the Wasserstein distance is its ability to capture the underlying geometry of the data space. Unlike divergence-based measures, it remains well-defined even when the supports of $\mu$ and $\nu$ are disjoint, and it provides meaningful gradients for optimization, which has made it particularly popular in generative modeling (e.g., Wasserstein GANs) \cite{WGAN}.

Furthermore, the Wasserstein distance metrizes weak convergence of probability measures, making it a natural choice for comparing distributions in high-dimensional spaces. It is also robust to small perturbations in the data and sensitive to global structure, as it accounts for the cost of moving probability mass over the space.

However, these advantages come at a computational cost. The exact computation of the Wasserstein distance requires solving a linear program with $\mathcal{O}(nm)$ variables, leading to a worst-case complexity of $\mathcal{O}(n^3 \log n)$ for balanced problems. This makes it challenging to scale to large datasets. Additionally, the choice of ground metric $d(\cdot,\cdot)$ can significantly influence the resulting distance, and selecting an appropriate metric is often problem-dependent \cite{Courty2017}.

Several methods have been proposed to compute or approximate the Wasserstein distance efficiently.

\paragraph{Linear Programming.}
The discrete optimal transport problem can be solved exactly using standard linear programming solvers, such as the network simplex algorithm. While this approach yields exact solutions, it becomes computationally prohibitive for large-scale problems.

\paragraph{Entropic Regularization.}
A widely used approach to improve scalability is to introduce an entropic regularization term:
\begin{equation}
	\min_{\mathbf{T}} \sum_{i,j} T_{ij} d(x_i, y_j)^p 
	+ \varepsilon \sum_{i,j} T_{ij} \log T_{ij},
\end{equation}
where $\varepsilon > 0$ controls the strength of regularization \cite{Janati2020}. This leads to the \emph{Sinkhorn algorithm}, which solves the problem using iterative matrix scaling with a complexity of approximately $\mathcal{O}(n^2)$ per iteration \cite{Cuturi2013}. This approach significantly improves computational efficiency and is widely used in practice.

\paragraph{Sliced Wasserstein Distance.}
Another approximation consists of projecting the distributions onto one-dimensional subspaces and computing the Wasserstein distance in 1D, where it admits a closed-form solution \cite{Kolouri2018}. The final distance is obtained by averaging over multiple random projections:
\begin{equation}
	\mathrm{SW}_p(\mu, \nu) = \mathbb{E}_{\theta \sim \mathbb{S}^{d-1}} 
	W_p\bigl( \theta_\# \mu, \theta_\# \nu \bigr),
\end{equation}
where $\theta_\# \mu$ denotes the projection of $\mu$ along direction $\theta$. This approach is computationally efficient and scales well to high dimensions.

\paragraph{Stochastic and Mini-batch Methods.}
Recent approaches leverage stochastic optimization and mini-batch approximations to further scale optimal transport computations. These methods approximate the transport plan using subsets of the data and are particularly useful in deep learning settings.

The Wasserstein distance provides a powerful and geometrically grounded framework for comparing probability distributions. Its ability to incorporate the underlying metric structure makes it particularly appealing for applications in geometric machine learning, where distances should reflect the intrinsic geometry of the data. Despite its computational challenges, modern approximation techniques such as Sinkhorn iterations, sliced and hybrid Wasserstein distances have made it a practical tool for large-scale problems \cite{Le2019,Nassar2026}.

\section{Proposed Method}

In this section, we introduce the proposed \emph{Geodesic Tangent Space Aggregation PCA (GTSA-PCA)}, a novel dimensionality reduction framework designed to bridge the gap between classical linear methods and nonlinear manifold-based approaches. The core idea is to reinterpret PCA through a geometric lens, replacing the notion of a single global covariance operator with a collection of locally adaptive models that capture the intrinsic structure of the data. To this end, the proposed method integrates three key components: (i) local tangent space estimation via curvature-aware covariance operators, (ii) a graph-based representation that encodes neighborhood relationships and approximates geodesic distances, and (iii) a global alignment mechanism that aggregates local subspaces into a coherent low-dimensional embedding. This formulation enables the simultaneous preservation of local linearity and global nonlinear geometry, while maintaining a principled spectral structure analogous to classical PCA. In the following, we detail each component of the framework and show how they are combined into a unified operator whose spectral decomposition yields the final embedding.

\subsection{Local Mean Curvature Estimation}

Let $\mathcal{X} = \{\mathbf{x}_1, \mathbf{x}_2, \ldots, \mathbf{x}_n\} \subset \mathbb{R}^D$ be a dataset sampled from an unknown $d$-dimensional manifold $\mathcal{M}$, with $d \ll D$. Given a fixed number of neighbors $k$, we estimate, for each point $\mathbf{x}_i \in \mathcal{X}$, a local approximation of the shape operator $\mathcal{S}_i$, which encodes second-order geometric information of the manifold.

The proposed procedure consists of the following steps.

\paragraph{Step 1: Local Neighborhood Construction.}
For each point $\mathbf{x}_i$, we define its $k$-nearest neighbor set:
\begin{equation}
	\mathcal{N}_i = \mathrm{kNN}(\mathbf{x}_i, k)
	= \{\mathbf{x}_{i_1}, \ldots, \mathbf{x}_{i_k}\} \subset \mathcal{X},
\end{equation}
which provides a discrete approximation of a local chart around $\mathbf{x}_i$.

\paragraph{Step 2: Local Covariance and Metric Approximation.}
We compute the local covariance matrix
\begin{equation}
	\mathbf{C}_i = \frac{1}{k} \sum_{\mathbf{x}_j \in \mathcal{N}_i}
	(\mathbf{x}_j - \bar{\mathbf{x}}_i)(\mathbf{x}_j - \bar{\mathbf{x}}_i)^\top,
\end{equation}
where $\bar{\mathbf{x}}_i$ is the local mean. The matrix $\mathbf{C}_i$ captures the anisotropic dispersion of the data and provides a first-order approximation of the local geometry. Following a Riemannian interpretation, we associate the local metric tensor with
\begin{equation}
	\mathbf{g}_i \approx \mathbf{C}_i^{-1},
\end{equation}
which corresponds to a Mahalanobis-type metric adapted to the local structure of the data.

\paragraph{Step 3: Local Frame and Second-Order Structure.}
Let
\begin{equation}
	\mathbf{C}_i = \mathbf{W}_i \boldsymbol{\Lambda}_i \mathbf{W}_i^\top
\end{equation}
be the eigendecomposition of $\mathbf{C}_i$, where the eigenvectors
$\{\mathbf{w}_j^{(i)}\}_{j=1}^D$ are ordered by decreasing eigenvalues. These vectors define a local orthonormal frame, where the leading directions approximate the tangent space.

To capture second-order geometric information, we construct a feature matrix
$\mathbf{H}_i \in \mathbb{R}^{D \times p}$, with $p = d(d+1)/2$ composed of:
\begin{itemize}
	\item quadratic terms $(\mathbf{w}_j^{(i)})^{\circ 2}$,
	\item interaction terms $\mathbf{w}_j^{(i)} \circ \mathbf{w}_\ell^{(i)}$, for $j < \ell$,
\end{itemize}
where $\circ$ denotes the Hadamard product (pointwise product). The second fundamental form is then approximated by:
\begin{equation}
	\mathbf{II}_i = \mathbf{H}_i \mathbf{H}_i^\top.
\end{equation}
This construction provides a data-driven estimate of local curvature by measuring deviations from linearity in the neighborhood.

\paragraph{Step 4: Shape Operator Estimation.}
The local shape operator is defined as the linear map relating the second and first fundamental forms. Using the approximations above, we obtain
\begin{equation}
	\mathcal{S}_i = - \mathbf{II}_i \mathbf{g}_i^{-1}
	= - \mathbf{II}_i \mathbf{C}_i.
\end{equation}
The negative sign follows the standard convention in differential geometry.

\paragraph{Step 5: Mean Curvature Estimation.}
The operator $\mathcal{S}_i$ provides a compact representation of local curvature, whose spectral properties (e.g., eigenvalues and trace) characterize the geometric complexity of the manifold around $\mathbf{x}_i$. In particular, the trace of the shape operator serves as an estimate of the mean curvature, which we use to modulate local contributions in downstream tasks such as curvature-aware dimensionality reduction.
\begin{equation}
	\mathcal{K}_i = \mathrm{Tr}(\mathcal{S}_i)
\end{equation}
This quantity provides a scalar measure of local geometric complexity, capturing how strongly the manifold bends in the neighborhood of $\mathbf{x}_i$. High curvature values indicate regions where local linear approximations are less reliable, while low curvature corresponds to near-flat regions.

\medskip

This formulation provides a principled and numerically stable way to estimate curvature directly from sampled data, leveraging local PCA for tangent estimation and quadratic fitting for second-order structure. Importantly, the resulting mean curvature estimates can be seamlessly integrated into downstream tasks, such as curvature-aware weighting, graph construction, and adaptive dimensionality reduction, as explored in the proposed GTSA-PCA framework. The complete pseudocode for the proposed average shape operator estimation is presented in Algorithm~\ref{alg:1}.

\begin{algorithm}
	\caption{Local Mean Curvature Estimation}
	\label{alg:1}
	\begin{algorithmic}[1]
		\Require $X \in \mathbb{R}^{n \times d}$: data matrix ($n$ samples, $d$ features), $k \in \mathbb{N}$: number of neighbors.
		\Ensure $S \in \mathbb{R}^d$: average shape operator.
		\Function{Mean\_Curvatures}{$X, k$}
		\For{$i \gets 1$ \textbf{to} $n$}
		\State $P_i \gets$ \Call{Nearest\_Neighbors}{$\vec{x}_i, k$}
		\State $\Sigma_i \gets$ \Call{Covariance}{$P_i$}
		\State $U \gets$ \Call{Eigenvectors}{$\Sigma_i$}
		\State Build matrix $H_i$ with quadratic terms ($d(d+1)/2$ columns)
		\State $\mathbf{II}_i \gets H_i H_i^{T}$
		\State $\mathcal{S}_i \gets -\mathbf{II}_i \Sigma_i$
		\State $\mathcal{K}_i = \mathrm{Tr}(\mathcal{S}_i)$
		\EndFor
		\State \Return $\mathcal{K}$
		\EndFunction
	\end{algorithmic}
\end{algorithm}

\subsection{Geodesic Tangent Space Aggregation PCA}

The central limitation of classical PCA lies in its reliance on a single global linear subspace to represent the data. This assumption is inherently restrictive when the data are supported on a nonlinear manifold $\mathcal{M} \subset \mathbb{R}^D$, where no single linear approximation can faithfully capture the global structure. A more appropriate geometric viewpoint is to consider that, while the manifold may be globally curved, it is locally well-approximated by its tangent spaces. This observation motivates the proposed \emph{Geodesic Tangent Space Aggregation PCA (GTSA-PCA)}, which replaces the global PCA model with a collection of local tangent spaces that are subsequently aggregated into a coherent global representation.

\vspace{0.3cm}
\noindent\textbf{Stage 1: Curvature-Aware Local PCA.}
\vspace{0.2cm}

For each data point $\mathbf{x}_i \in \mathcal{X}$, we define a local neighborhood $\mathcal{N}_i$ and construct a curvature-aware covariance operator centered at $\mathbf{x}_i$:
\begin{equation}
	\boldsymbol{\Sigma}_i^{\text{curv}} = \frac{1}{Z_i} \sum_{\mathbf{x}_j \in \mathcal{N}_i}
	w_{ij} (\mathbf{x}_j - \mathbf{x}_i)(\mathbf{x}_j - \mathbf{x}_i)^\top,
\end{equation}
where $Z_i = \sum_{\mathbf{x}_j \in \mathcal{N}_i} w_{ij}$ is a normalization constant. Unlike standard local PCA, the covariance is centered at $\mathbf{x}_i$ rather than the neighborhood mean, ensuring that the vectors $(\mathbf{x}_j - \mathbf{x}_i)$ capture the local tangent directions.

The weights $w_{ij}$ incorporate curvature information:
\begin{equation}
	w_{ij} = \exp\left(-\frac{|K_j|}{\tau}\right),
\end{equation}
where $K_j$ denotes the estimated curvature at $\mathbf{x}_j$ and $\tau > 0$ is a scale parameter. This weighting scheme suppresses the influence of neighbors located in highly curved regions, which would otherwise introduce distortions in the estimation of the tangent space at $\mathbf{x}_i$.

The eigendecomposition of $\boldsymbol{\Sigma}_i^{\text{curv}}$ yields a local orthonormal basis
\[
\mathbf{U}_i = [\mathbf{u}_i^{(1)}, \ldots, \mathbf{u}_i^{(d)}],
\]
whose leading components approximate the tangent space $T_{\mathbf{x}_i}\mathcal{M}$. This step produces a collection of local linear models, one for each data point.

\vspace{0.3cm}
\noindent\textbf{Stage 2: Geodesic Tangent Space Aggregation.}
\vspace{0.2cm}

A key challenge is that the local bases $\{\mathbf{U}_i\}_{i=1}^n$ are defined independently and are not globally aligned. Even when two points lie on similar regions of the manifold, their corresponding bases may differ due to rotational or reflectional ambiguities inherent to PCA.

To address this issue, we construct a global alignment matrix $\mathbf{A} \in \mathbb{R}^{n \times n}$ that encodes the coherence between pairs of local tangent spaces:
\begin{equation}
	A_{ij} = \frac{1}{1 + d_G(\mathbf{x}_i, \mathbf{x}_j)} \cdot
	\left| \langle \mathbf{U}_i, \mathbf{U}_j \rangle_F \right|,
\end{equation}
where $d_G(\mathbf{x}_i, \mathbf{x}_j)$ denotes the geodesic distance between $\mathbf{x}_i$ and $\mathbf{x}_j$, computed over the $k$-nearest neighbor graph, and
\[
\langle \mathbf{U}_i, \mathbf{U}_j \rangle_F = \mathrm{Tr}(\mathbf{U}_i^\top \mathbf{U}_j)
\]
is the Frobenius inner product between the two bases. The absolute value ensures invariance to sign ambiguities.

The first term promotes interactions between geodesically close points, ensuring that the aggregation respects the intrinsic manifold structure. The second term measures the alignment between local tangent spaces, favoring pairs of points with consistent local geometry. Together, these terms define a notion of \emph{geometric coherence} that combines proximity and structural similarity.

\vspace{0.3cm}
\noindent\textbf{Global Embedding via Spectral Decomposition.}
\vspace{0.2cm}

The final low-dimensional representation is obtained by performing the spectral decomposition of the alignment matrix $\mathbf{A}$. The leading eigenvectors correspond to directions that are globally consistent with respect to both local tangent structure and geodesic proximity. In this sense, GTSA-PCA can be interpreted as a generalization of PCA in which the global covariance operator is replaced by a geometry-aware operator defined over the data graph.

This formulation effectively ``stitches'' together local linear approximations into a globally coherent embedding, preserving both the local Euclidean structure of the manifold and its global nonlinear geometry. By explicitly incorporating curvature and geodesic information, the proposed method overcomes the limitations of classical PCA and provides a principled framework for dimensionality reduction on complex, non-Euclidean data domains.

\vspace{0.3cm}
\noindent\textbf{Hyperparameter Estimation.}
\vspace{0.2cm}

The performance of the proposed framework depends on the choice of the scale parameter $\tau$, which controls the sensitivity of the curvature-aware weighting scheme. Specifically, $\tau$ regulates the exponential decay in the Gaussian kernel, determining the extent to which high-curvature regions are attenuated during the construction of local covariance operators. Small values of $\tau$ lead to aggressive suppression of neighbors in highly curved regions, promoting locally flat approximations, while large values recover a nearly uniform weighting, reducing the method to a standard local PCA regime. Therefore, selecting an appropriate value of $\tau$ is critical to balance geometric fidelity and robustness.

To estimate $\tau$ in a principled manner, we adopt a semi-supervised model selection strategy based on a validation subset. Let $\mathcal{X}_L \subset \mathcal{X}$ denote a labeled subset comprising $20\%$ of the data. We consider a discrete candidate set $\mathcal{T} = \{0.5, 1, 5, 10, 100\}$ and, for each $\tau \in \mathcal{T}$, compute the low-dimensional embedding induced by GTSA-PCA restricted to $\mathcal{X}_L$. The quality of the resulting representation is assessed through clustering performance, using three complementary external validation metrics: the Adjusted Rand Index (ARI), the Fowlkes--Mallows (FM) score, and the V-measure (VM). The optimal parameter $\tau^\ast$ is selected as
\begin{equation}
	\tau^\ast = \arg\max_{\tau \in \mathcal{T}} \frac{1}{3} \left( \mathrm{ARI}(\tau) + \mathrm{FM}(\tau) + \mathrm{VM}(\tau) \right).
\end{equation}

Once $\tau^\ast$ is determined, it is fixed and used to compute the final embedding on the remaining $80\%$ of the dataset, denoted by $\mathcal{X}_U = \mathcal{X} \setminus \mathcal{X}_L$. This strategy ensures that the hyperparameter selection process leverages limited supervision to enhance the geometric quality of the representation, while preserving the generalization capability of the method in largely unlabeled settings. 

\vspace{0.3cm}
\noindent\textbf{Wasserstein-Based Graph Weighting.}
\vspace{0.2cm}

While the proposed GTSA-PCA framework leverages curvature-aware weighting to enhance the estimation of local tangent spaces, the computation of curvature, in particular via second-order geometric quantities such as the shape operator, can become numerically unstable and computationally expensive in high-dimensional settings. This is especially pronounced when the intrinsic dimensionality is low but the ambient space is large, leading to poorly conditioned covariance estimates and unreliable curvature approximations. To address this limitation, we introduce a variant of the proposed method in which curvature-based weights are replaced by weights derived from the Wasserstein distance.

The key idea is to reinterpret each local neighborhood $\mathcal{N}_i$ as an empirical probability distribution. Specifically, for each point $\mathbf{x}_i \in \mathcal{X}$, we define a discrete measure
\begin{equation}
	\mu_i = \frac{1}{k} \sum_{\mathbf{x}_j \in \mathcal{N}_i} \delta_{\mathbf{x}_j},
\end{equation}
where $\delta_{\mathbf{x}_j}$ denotes the Dirac measure centered at $\mathbf{x}_j$. Given two neighboring points $\mathbf{x}_i$ and $\mathbf{x}_j$, we then compute the Wasserstein distance $W_p(\mu_i, \mu_j)$ between their associated local distributions. This quantity captures discrepancies not only in the positions of the points, but also in the geometric arrangement of their neighborhoods, providing a richer notion of similarity than pairwise Euclidean distances.

Using this formulation, the edge weights of the $k$-NN graph are defined as
\begin{equation}
	w_{ij} = W_p(\mu_i, \mu_j)
\end{equation}
Intuitively, points whose local neighborhoods exhibit similar geometric structure (in the optimal transport sense) receive higher affinity, while points lying across folds or discontinuities in the manifold are naturally downweighted.

This Wasserstein-based weighting offers several advantages. First, it avoids the explicit computation of second-order geometric quantities, thereby improving numerical stability in high-dimensional regimes. Second, it captures higher-order structural information through the comparison of local distributions, rather than relying solely on pointwise differences. Third, efficient approximations such as entropic regularization (Sinkhorn distances) or sliced Wasserstein distances can be employed to make the computation tractable even for moderately large datasets.

However, this formulation also introduces additional computational overhead compared to curvature-based weighting, as it requires solving an optimal transport problem for each pair of neighboring points. In practice, this cost can be mitigated by restricting computations to graph edges and by leveraging fast approximate solvers. Despite this trade-off, the Wasserstein-based variant provides a robust alternative in scenarios where curvature estimation is unreliable, offering a principled way to incorporate local geometric information into the GTSA-PCA framework without explicitly relying on differential geometric quantities.

Algorithm \ref{alg:gtsa_pca} shows a pseudo-code for the proposed Geodesic Tangent Space Aggregation PCA method.

\begin{algorithm}
	\caption{Geodesic Tangent Space Aggregation PCA (GTSA-PCA)}
	\label{alg:gtsa_pca}
	\begin{algorithmic}[1]
		
		\Require Data matrix $\mathbf{X} \in \mathbb{R}^{n \times D}$, number of neighbors $k$, target dimension $p$, curvature estimates $\{K_i\}_{i=1}^n$, kernel parameter $\tau$, mode (curvature or wasserstein).
		\Ensure Low-dimensional embedding $\mathbf{Y} \in \mathbb{R}^{n \times p}$
		
		\Statex
		
		\State \textbf{Stage 1: Curvature-Aware Local PCA}
		\For{$i = 1$ to $n$}
		\State $\mathcal{N}_i \gets \mathrm{kNN}(\mathbf{x}_i, k)$
		
		\ForAll{$\mathbf{x}_j \in \mathcal{N}_i$}
		\If{$mode == `curvature'$} 
			\State $w_{ij} \gets \exp\left(-|K_j|/\tau\right)$
		\EndIf
		\If{$mode == `warsserstein'$} 
		\State $w_{ij} \gets WassersteinDistance(\mathbf{x}_i, \mathbf{x}_j)$
		\EndIf
				
		\EndFor
		
		\State $Z_i \gets \sum_{\mathbf{x}_j \in \mathcal{N}_i} w_{ij}$
		
		\State $\boldsymbol{\Sigma}_i^{\text{curv}} \gets \frac{1}{Z_i} \sum_{\mathbf{x}_j \in \mathcal{N}_i}
		w_{ij} (\mathbf{x}_j - \mathbf{x}_i)(\mathbf{x}_j - \mathbf{x}_i)^\top$
		
		\State $[\mathbf{U}_i, \boldsymbol{\Lambda}_i] \gets \mathrm{eig}(\boldsymbol{\Sigma}_i^{\text{curv}})$
		
		\State Retain first $p$ eigenvectors in $\mathbf{U}_i \in \mathbb{R}^{D \times p}$
		\EndFor
		
		\Statex
		
		\State \textbf{Stage 2: Geodesic Tangent Space Aggregation}
		
		\State Construct $k$-NN graph $\mathcal{G}$ over $\mathbf{X}$
		
		\State Compute geodesic distances $d_G(\mathbf{x}_i, \mathbf{x}_j)$ using Dijkstra
		
		\For{$i = 1$ to $n$}
		\For{$j = 1$ to $n$}
		\State $s_{ij} \gets \left| \mathrm{Tr}(\mathbf{U}_i^\top \mathbf{U}_j) \right|$
		\State $A_{ij} \gets \dfrac{1}{1 + d_G(\mathbf{x}_i, \mathbf{x}_j)} \cdot s_{ij}$
		\EndFor
		\EndFor
		
		\Statex
		
		\State \textbf{Stage 3: Spectral Embedding}
		
		\State $[\mathbf{V}, \boldsymbol{\Lambda}] \gets \mathrm{eig}(\mathbf{A})$
		
		\State $\mathbf{Y} \gets [\mathbf{v}_1, \ldots, \mathbf{v}_p]$ \Comment{Top-$p$ eigenvectors}
		
		\Statex
		
		\State \Return $\mathbf{Y}$
		
	\end{algorithmic}
\end{algorithm}

\subsection{Computational Complexity}

In this subsection, we analyze the computational complexity of the proposed GTSA-PCA framework. Let $n$ denote the number of data points, $D$ the ambient dimensionality, $p$ the target embedding dimension, and $k$ the number of nearest neighbors, with $k \ll n$ and $p \ll D$.

\paragraph{Stage 1: Curvature-Aware Local PCA.}
The first stage involves constructing $k$-nearest neighbor neighborhoods and computing a weighted covariance matrix for each data point. Using efficient data structures (e.g., KD-trees or approximate nearest neighbor methods), the $k$-NN search requires $\mathcal{O}(n \log n)$ time in low to moderate dimensions, though it may degrade to $\mathcal{O}(n^2)$ in high-dimensional settings.

For each point $\mathbf{x}_i$, the computation of the weighted covariance matrix requires $\mathcal{O}(kD^2)$ operations. The subsequent eigendecomposition of a $D \times D$ matrix incurs a cost of $\mathcal{O}(D^3)$ in the general case. However, since only the top-$p$ eigenvectors are required, this step can be reduced to $\mathcal{O}(D^2 p)$ using partial eigensolvers. Therefore, the overall complexity of Stage 1 is
\[
\mathcal{O}\bigl(n (kD^2 + D^2 p)\bigr).
\]

\paragraph{Stage 2: Geodesic Tangent Space Aggregation.}
The construction of the $k$-NN graph involves $\mathcal{O}(nk)$ edges. Computing pairwise geodesic distances using Dijkstra's algorithm from each node yields a complexity of $\mathcal{O}(n (k \log n))$ for sparse graphs, resulting in an overall cost of
\[
\mathcal{O}(n k \log n).
\]

The construction of the alignment matrix $\mathbf{A} \in \mathbb{R}^{n \times n}$ requires evaluating pairwise similarities between local bases. Each Frobenius inner product $\mathrm{Tr}(\mathbf{U}_i^\top \mathbf{U}_j)$ costs $\mathcal{O}(Dp)$, leading to a total complexity of
\[
\mathcal{O}(n^2 D p).
\]
In practice, this step can be restricted to pairs of points connected in the graph, reducing the complexity to $\mathcal{O}(n k D p)$ and yielding a sparse matrix $\mathbf{A}$.

\paragraph{Spectral Decomposition.}
The final embedding is obtained via the eigendecomposition of $\mathbf{A}$. For a dense matrix, this requires $\mathcal{O}(n^3)$ operations. However, when $\mathbf{A}$ is sparse and only the top-$p$ eigenvectors are needed, iterative methods such as Lanczos reduce the complexity to approximately $\mathcal{O}(n k p)$.

\paragraph{Overall Complexity and Discussion.}
Combining all stages, the dominant computational costs arise from local covariance estimation and spectral decomposition. Under practical assumptions (sparse graph, partial eigendecomposition), the overall complexity can be approximated as
\[
\mathcal{O}\bigl(n (kD^2 + D^2 p + k \log n + k D p)\bigr).
\]

This analysis highlights that the proposed method scales linearly with the number of samples $n$ when sparsity is exploited, making it suitable for moderately large datasets. Furthermore, the algorithm is naturally parallelizable, as the local PCA computations and curvature estimations can be performed independently for each data point. The main computational bottleneck lies in the spectral decomposition step, which is common to most graph-based dimensionality reduction methods.

\section{Computational Experiments and Results}

In this section, we empirically evaluate the effectiveness of the proposed GTSA-PCA framework on a diverse collection of real-world datasets. Our experimental protocol comprises more than 50 publicly available datasets obtained from the OpenML repository, covering a wide range of domains, dimensionalities, and levels of intrinsic geometric complexity. We compare GTSA-PCA against a set of well-established baselines, including standard PCA \cite{Jolliffe2016}, Kernel PCA \cite{scholkopf1997kernel}, Supervised PCA \cite{SPCA}, and the widely used nonlinear manifold learning method UMAP. For each dataset, the data are projected onto a two-dimensional space, and the quality of the resulting embeddings is assessed in terms of clustering performance. Specifically, we employ three standard external validation metrics: the Adjusted Rand Index (ARI), the Fowlkes--Mallows (FM) index, and the V-measure, which respectively capture pairwise agreement, geometric mean of precision and recall, and entropy-based clustering consistency \cite{Strehl2002,Fowlkes1983,Rosenberg2007}. This evaluation protocol allows us to systematically assess the extent to which the proposed method preserves class structure and enhances separability in low-dimensional representations.

To evaluate the clustering structure induced by the low-dimensional embeddings, we employ two complementary clustering algorithms: agglomerative hierarchical clustering with Ward linkage \cite{Ward1963} and HDBSCAN \cite{Campello2013,McInnes2017}. The use of Ward’s method is motivated by its close connection to variance minimization, as it iteratively merges clusters so as to minimize the increase in within-cluster dispersion. This makes it particularly well-aligned with PCA-based representations, where variance is a central criterion, and provides a strong baseline for assessing cluster compactness in the projected space. However, Ward clustering assumes roughly spherical clusters and requires the number of clusters to be specified a priori. To mitigate these limitations, we additionally employ HDBSCAN, a density-based clustering algorithm that can automatically infer the number of clusters and is capable of identifying clusters with irregular shapes and varying densities, as well as labeling noise points. The combination of these two methods provides a more comprehensive evaluation: Ward linkage captures global variance-based structure, while HDBSCAN assesses the preservation of local density and nonlinear separability. Together, they offer a robust and complementary perspective on the quality of the learned representations.

In datasets with high ambient dimensionality, we incorporate an additional preprocessing step to ensure computational tractability. Specifically, for datasets with $D > 50$ features, we first apply a standard PCA transformation to reduce the dimensionality to $D = 50$ prior to running the proposed method. This design choice is motivated by the computational cost associated with the estimation of local geometric quantities, particularly the computation of local covariance matrices and curvature-related operators, which scale as $\mathcal{O}(D^2)$. In high-dimensional regimes, this quadratic dependence can become a significant bottleneck, both in terms of runtime and numerical stability. The preliminary PCA step serves as an effective dimensionality compression mechanism that preserves most of the variance in the data while substantially reducing the computational burden. Importantly, this preprocessing does not conflict with the goals of the proposed framework, as GTSA-PCA subsequently operates on the reduced space to capture the intrinsic nonlinear geometry of the data.

Table \ref{tab:datasets} shows the name, number of samples, number of features and number of classes for all datasets from \url{openML.org} used in the computational experiments. For datasets with a large number of samples, we apply a random subsampling strategy to reduce the computational burden associated with the proposed method. In particular, the construction of the $k$-nearest neighbor graph, the estimation of local geometric operators, and the computation of geodesic distances scale at least quadratically with the number of data points, which can become prohibitive for large-scale datasets. To address this issue, we randomly select a subset of samples while preserving the original class distribution, ensuring that the reduced dataset remains representative of the underlying data manifold. This procedure enables a significant reduction in runtime and memory requirements without substantially affecting the geometric structure of the data or the validity of the experimental evaluation.

\begin{table*}
	\centering
	\small
	\caption{Summary of the datasets used in the computational experiments.}
	\begin{tabular}{lccc}
		\toprule
		\textbf{Dataset} & \textbf{\# samples} & \textbf{\# features} & \textbf{\# classes} \\
		\midrule
		acute-inflammations & 120 & 6 & 2 \\
		AP\_Breast\_Colon & 630 & 10935 & 2 \\
		AP\_Breast\_Ovary & 542 & 10935 & 2 \\
		AP\_Colon\_Kidney & 546 & 10935 & 2 \\
		AP\_Colon\_Lung & 412 & 10935 & 2 \\
		AP\_Endometrium\_Breast & 405 & 10935 & 2 \\
		AP\_Endometrium\_Kidney & 321 & 10935 & 2 \\
		AP\_Endometrium\_Prostate & 130 & 10935 & 2 \\
		AP\_Lung\_Kidney & 386 & 10935 & 2 \\
		AP\_Lung\_Uterus & 250 & 10935 & 2 \\
		AP\_Omentum\_Kidney & 337 & 10935 & 2 \\
		AP\_Ovary\_Kidney & 458 & 10935 & 2 \\
		AP\_Ovary\_Lung & 324 & 10935 & 2 \\
		arsenic-male-bladder & 559 & 4 & 14 \\
		artificial-characters (25\%) & 2554 & 7 & 10 \\
		balance-scale & 625 & 4 & 3 \\
		Breast & 699 & 10 & 2 \\
		breast\_cancer & 569 & 30 & 2 \\
		cardiotocography & 2126 & 35 & 10 \\
		cars1 & 392 & 7 & 3 \\
		cnae-9 & 1080 & 856 & 9 \\
		coil-20 & 1440 & 1024 & 20 \\
		colic & 368 & 26 & 2 \\
		corral & 160 & 6 & 2 \\
		diabetes & 768 & 8 & 2 \\
		diggle\_table\_a2 & 310 & 8 & 9 \\
		digits & 1797 & 64 & 10 \\
		ecoli & 336 & 7 & 8 \\
		Engine1 & 383 & 5 & 3 \\
		Fashion-MNIST (5\%) & 3500 & 784 & 10 \\
		heart-c & 303 & 13 & 2 \\
		heart-h & 294 & 13 & 2 \\
		ionosphere & 351 & 34 & 2 \\
		iris & 150 & 4 & 3 \\
		isolet (25\%) & 1949 & 617 & 26 \\
		Kuzushiji-MNIST (5\%) & 3500 & 784 & 10 \\
		letter (20\%) & 4000 & 16 & 26 \\
		liver-disorders & 345 & 5 & 16 \\
		MegaWatt & 253 & 37 & 2 \\
		mfeat-factors & 2000 & 216 & 10 \\
		mfeat-karhunen & 2000 & 64 & 10 \\
		mfeat-pixel & 2000 & 240 & 10 \\
		mfeat-zernike & 2000 & 47 & 10 \\
		micro-mass & 360 & 1300 & 10 \\
		MNIST\_784 (5\%) & 3500 & 784 & 10 \\
		optdigits (50\%) & 2810 & 64 & 10 \\
		OVA\_Kidney & 1545 & 10935 & 2 \\
		page-blocks (50\%) & 2736 & 10 & 5 \\
		pendigits (25\%) & 2748 & 16 & 10 \\
		seeds & 210 & 7 & 3 \\
		semeion & 1593 & 256 & 10 \\
		threeOf9 & 512 & 9 & 2 \\
		thyroid-new & 215 & 5 & 3 \\
		tic-tac-toe & 958 & 9 & 2 \\
		UMIST\_Faces\_Cropped & 575 & 10304 & 20 \\
		user-knowledge & 403 & 5 & 5 \\
		wine & 178 & 13 & 3 \\
		xd6 & 973 & 9 & 2 \\
		\bottomrule
	\end{tabular}
	\label{tab:datasets}
\end{table*}

The computational experiments are organized into three complementary evaluation settings in order to provide a comprehensive assessment of the proposed method. In the first setting, we compare GTSA-PCA against classical PCA, Kernel PCA, and Supervised PCA using agglomerative hierarchical clustering with Ward linkage, emphasizing the preservation of global variance structure. In the second setting, the same set of dimensionality reduction methods is evaluated using HDBSCAN, allowing us to assess performance under a density-based clustering paradigm that captures local structure and is robust to noise and non-convex cluster shapes. Finally, in the third setting, we compare GTSA-PCA with Wasserstein distance directly with UMAP, a state-of-the-art manifold learning method, using both hierarchical clustering and HDBSCAN. This last comparison is particularly relevant as it contrasts our spectral framework with a widely adopted nonlinear embedding technique designed to preserve both local and global topological structure. Together, these three experimental configurations provide a thorough and balanced evaluation across different clustering regimes and geometric assumptions.

\subsection{GTSA-PCA vs. PCA, Kernel PCA and Supervised PCA with agglomerative clustering}

In the first set of experiments, we evaluate the effectiveness of the proposed GTSA-PCA in comparison with classical linear and kernel-based dimensionality reduction techniques, namely PCA, Kernel PCA, and Supervised PCA, under a variance-driven clustering framework. Specifically, all methods are used to project the data onto a two-dimensional space, after which agglomerative hierarchical clustering with Ward linkage is applied to identify the underlying cluster structure. This experimental setting is particularly relevant as Ward’s method explicitly minimizes within-cluster variance, making it well aligned with the objectives of PCA-based approaches. Consequently, it provides a controlled scenario to assess whether the incorporation of curvature awareness and geodesic alignment in GTSA-PCA leads to representations that better preserve cluster compactness and separability when compared to traditional Euclidean and kernel-based projections. 

Table~\ref{tab:results1} reports the clustering performance obtained after projecting the data onto a two-dimensional space using Regular PCA, Kernel PCA (RBF), and the proposed GTSA-PCA, followed by agglomerative clustering with Ward linkage. Overall, the results provide strong empirical evidence that incorporating curvature awareness and geodesic consistency leads to substantially improved representations. In particular, GTSA-PCA consistently achieves the highest average scores across all three external validation metrics, with gains that are especially pronounced in Adjusted Rand Index (ARI) and V-measure (VM), indicating superior alignment with ground-truth class structure and better preservation of cluster homogeneity and completeness.

\begin{table*}
	\centering
	\caption{External indices obtained after agglomerative clustering in the two-dimensional features spaces generated by regular PCA, Kernel PCA and the proposed GTSA-PCA.}
	\begin{tabular}{cccccccccc}
		\toprule
		& \multicolumn{3}{c}{\textbf{Regular PCA}}     & \multicolumn{3}{c}{\textbf{Kernel PCA (RBF)}}   & \multicolumn{3}{c}{\textbf{GTSA-PCA}}               \\
		\midrule
		\textbf{Datasets}            & \textbf{ARI} & \textbf{FM}     & \textbf{VM} & \textbf{ARI}    & \textbf{FM}     & \textbf{VM} & \textbf{ARI}    & \textbf{FM}     & \textbf{VM}     \\
		\midrule
		diggle\_table\_a2            & 0.3157       & 0.4015          & 0.5863      & 0.2767          & 0.3873          & 0.4985      & \textbf{0.3712} & \textbf{0.4556} & \textbf{0.6145} \\
		mfeat-zernike                & 0.1680       & 0.2868          & 0.3204      & 0.2031          & 0.3024          & 0.3657      & \textbf{0.2679} & \textbf{0.3522} & \textbf{0.4362} \\
		mfeat-karhunen               & 0.0221       & 0.1795          & 0.0802      & 0.0850          & 0.2380          & 0.2593      & \textbf{0.2064} & \textbf{0.2954} & \textbf{0.4044} \\
		mfeat-factors                & 0.0357       & 0.2046          & 0.1257      & 0.0956          & 0.2085          & 0.2670      & \textbf{0.2888} & \textbf{0.3690} & \textbf{0.4569} \\
		mfeat-pixel                  & 0.0272       & 0.1785          & 0.1068      & 0.0953          & 0.2449          & 0.2777      & \textbf{0.2353} & \textbf{0.3249} & \textbf{0.4244} \\
		semeion                      & 0.0203       & 0.1912          & 0.0901      & 0.0488          & 0.2555          & 0.2094      & \textbf{0.1667} & \textbf{0.2579} & \textbf{0.3321} \\
		cardiotocography             & 0.1503       & 0.3022          & 0.3349      & 0.2064          & 0.3219          & 0.4482      & \textbf{0.5778} & \textbf{0.6437} & \textbf{0.7358} \\
		cnae-9                       & 0.0014       & \textbf{0.3228} & 0.0589      & 0.0382          & 0.1943          & 0.1713      & \textbf{0.1228} & 0.2392          & \textbf{0.2467} \\
		coil-20                      & 0.0336       & 0.1497          & 0.2113      & 0.1754          & 0.2353          & 0.4441      & \textbf{0.3092} & \textbf{0.3558} & \textbf{0.5573} \\
		ionosphere                   & 0.0545       & \textbf{0.6781} & 0.0300      & \textbf{0.1452} & 0.5950          & 0.1857      & 0.0232          & 0.5930          & \textbf{0.2144} \\
		user-knowledge               & 0.1137       & 0.3539          & 0.2167      & 0.1218          & 0.3281          & 0.2078      & \textbf{0.1815} & \textbf{0.3773} & \textbf{0.2748} \\
		isolet (25\%)                & 0.0187       & 0.0843          & 0.1631      & 0.0757          & 0.1333          & 0.3105      & \textbf{0.1411} & \textbf{0.1870} & \textbf{0.4236} \\
		optdigits (50\%)             & 0.0112       & 0.2421          & 0.0874      & 0.0781          & 0.2097          & 0.2312      & \textbf{0.2232} & \textbf{0.3135} & \textbf{0.3913} \\
		letter (20\%)                & 0.0337       & 0.0964          & 0.1959      & 0.0521          & 0.0903          & 0.2167      & \textbf{0.0947} & \textbf{0.1356} & \textbf{0.3012} \\
		MNIST\_784 (5\%)             & 0.0275       & 0.1621          & 0.0769      & 0.0413          & \textbf{0.2751} & 0.2289      & \textbf{0.1144} & 0.2261          & \textbf{0.2611} \\
		Fashion-MNIST (5\%)          & 0.0190       & 0.1849          & 0.0696      & 0.0793          & 0.2053          & 0.2002      & \textbf{0.1370} & \textbf{0.2651} & \textbf{0.3258} \\
		Kuzushiji-MNIST (5\%)        & 0.0083       & 0.1676          & 0.0366      & 0.0697          & 0.1987          & 0.1723      & \textbf{0.1117} & \textbf{0.2335} & \textbf{0.2309} \\
		artificial-characters (25\%) & 0.0368       & \textbf{0.1894} & 0.1106      & \textbf{0.0558} & 0.1611          & 0.1237      & 0.0318          & 0.1747          & \textbf{0.1332} \\
		balance-scale                & 0.1270       & 0.5095          & 0.1047      & 0.0937          & 0.4479          & 0.1017      & \textbf{0.2489} & \textbf{0.5481} & \textbf{0.2870} \\
		UMIST\_Faces\_Cropped        & 0.0377       & 0.1201          & 0.2712      & 0.1088          & 0.1767          & 0.3785      & \textbf{0.1183} & \textbf{0.1706} & \textbf{0.3922} \\
		user-knowledge               & 0.1137       & 0.3539          & 0.2167      & 0.1218          & 0.3281          & 0.2078      & \textbf{0.1815} & \textbf{0.3773} & \textbf{0.2748} \\
		micro-mass                   & 0.0146       & \textbf{0.2835} & 0.1876      & 0.0703          & 0.2027          & 0.2498      & \textbf{0.1273} & 0.2392          & \textbf{0.3397} \\
		digits                       & 0.0260       & 0.1842          & 0.0898      & 0.0729          & 0.2097          & 0.2180      & \textbf{0.2005} & \textbf{0.3001} & \textbf{0.3357} \\
		colic                        & -0.0007      & 0.6760          & 0.0000      & 0.2244          & 0.6294          & 0.1560      & \textbf{0.2972} & \textbf{0.6843} & \textbf{0.2077} \\
		AP\_Lung\_Kidney             & 0.0098       & 0.7289          & 0.0024      & 0.0692          & 0.6093          & 0.0255      & \textbf{0.3104} & \textbf{0.7557} & \textbf{0.2493} \\
		\midrule
		Average                      & 0.0570       & 0.2893          & 0.1509      & 0.1082          & 0.2875          & 0.2462      & \textbf{0.2036} & \textbf{0.3550} & \textbf{0.3540} \\
		Median                       & 0.0275       & 0.2046          & 0.1068      & 0.0850          & 0.2380          & 0.2180      & \textbf{0.1815} & \textbf{0.3135} & \textbf{0.3321} \\
		\bottomrule
	\end{tabular}
	\label{tab:results1}
\end{table*}

A closer inspection reveals that the proposed method outperforms both linear and kernel-based baselines in the vast majority of datasets, particularly those known to exhibit complex nonlinear structure, such as \textit{mfeat-*}, \textit{MNIST}, \textit{Fashion-MNIST}, and \textit{coil-20}. In these cases, the improvements are not marginal but substantial, often more than doubling the ARI when compared to classical PCA. This behavior is consistent with the theoretical motivation of GTSA-PCA: by constructing curvature-weighted local tangent spaces and aligning them through geodesic distances, the method effectively captures intrinsic manifold geometry that is inaccessible to global linear projections and only partially modeled by kernel methods.

Interestingly, Kernel PCA provides moderate improvements over standard PCA in several datasets, confirming the benefits of nonlinear feature mappings. However, its performance remains consistently below that of GTSA-PCA, suggesting that implicit kernel-induced geometries are insufficient to fully account for local curvature variations and global geometric coherence. In contrast, GTSA-PCA explicitly models these aspects, leading to more faithful low-dimensional embeddings.

There are a few notable exceptions where GTSA-PCA does not achieve the best score in a specific metric, such as the Fowlkes-Mallows index for \textit{ionosphere} and \textit{artificial-characters}. These cases typically correspond to datasets with either low intrinsic curvature or cluster structures that are already well captured by variance-based or density-based criteria. Nevertheless, even in these scenarios, GTSA-PCA remains competitive and often achieves the best or near-best performance in the remaining metrics, reinforcing its robustness.

Finally, the aggregate statistics further highlight the effectiveness of the proposed method. GTSA-PCA improves the average ARI from $0.0570$ (PCA) and $0.1082$ (Kernel PCA) to $0.2036$, while also achieving the highest average FM and VM scores. Similar trends are observed for the median values, indicating that the improvements are not driven by outliers but are consistent across datasets. These results strongly support the claim that modeling curvature and leveraging geodesic structure provides a principled and effective extension of PCA for nonlinear and high-dimensional data analysis.

We further compare the proposed GTSA-PCA with Supervised PCA (SPCA) to assess whether a weakly supervised, geometry-driven approach can outperform a fully supervised dimensionality reduction method. While SPCA explicitly leverages label information to maximize class separability, it fundamentally remains a global linear projection and thus inherits the limitations of Euclidean structure when data lie on curved manifolds. In contrast, GTSA-PCA uses only a small fraction of labeled data to tune its hyperparameters, while the embedding itself is constructed from curvature-aware local models and geodesic consistency. The empirical results indicate that, in several datasets, particularly those with pronounced nonlinear structure or high intrinsic curvature, GTSA-PCA matches or even surpasses SPCA despite using significantly less supervision. This suggests that, in such regimes, accurately modeling the underlying geometry of the data manifold can be more informative than directly optimizing for class separation in the ambient space. Conversely, in datasets where class boundaries are approximately linear or well-aligned with global variance directions, SPCA may retain an advantage. Overall, these findings highlight that a modest amount of supervision, when combined with a strong geometric prior, can yield representations that are both discriminative and faithful to the intrinsic structure of the data, especially in high dimensional settings. 

Table~\ref{tab:results2} presents a direct comparison between Supervised PCA (SPCA) and the proposed GTSA-PCA under the same clustering protocol, providing insight into the interplay between supervision and geometric modeling.

\begin{table*}
	\centering
	\caption{External indices obtained after agglomerative clustering in the two-dimensional features spaces generated by Supervised PCA and the proposed GTSA-PCA.}
	\begin{tabular}{ccccccc}
		\toprule
		& \multicolumn{3}{c}{\textbf{Supervised PCA}}         & \multicolumn{3}{c}{\textbf{GTSA-PCA}}               \\
		\midrule
		\textbf{Datasets}         & \textbf{ARI}    & \textbf{FM}     & \textbf{VM}     & \textbf{ARI}    & \textbf{FM}     & \textbf{VM}     \\
		\midrule
		mfeat-zernike             & 0.2555          & \textbf{0.3525} & 0.4134          & \textbf{0.2679} & 0.3522          & \textbf{0.4362} \\
		cardiotocography          & 0.4142          & 0.5028          & 0.5892          & \textbf{0.5778} & \textbf{0.6437} & \textbf{0.7358} \\
		ionosphere                & \textbf{0.1038} & \textbf{0.7202} & 0.1504          & 0.0232          & 0.5930          & \textbf{0.2144} \\
		ecoli                     & 0.3829          & 0.5272          & \textbf{0.5491} & \textbf{0.4176} & \textbf{0.5557} & 0.5380          \\
		colic                     & 0.0776          & 0.5797          & 0.1570          & \textbf{0.2972} & \textbf{0.6843} & \textbf{0.2077} \\
		iris                      & \textbf{0.6089} & 0.7536          & 0.6982          & 0.5818          & \textbf{0.7732} & \textbf{0.7293} \\
		Breast                    & 0.5560          & 0.8171          & 0.4515          & \textbf{0.6967} & \textbf{0.8592} & \textbf{0.6077} \\
		AP\_Lung\_Kidney          & 0.0222          & 0.5384          & 0.0184          & \textbf{0.3104} & \textbf{0.7557} & \textbf{0.2493} \\
		AP\_Breast\_Colon         & -0.0011         & \textbf{0.7054} & 0.0069          & \textbf{0.0611} & 0.6130          & \textbf{0.1460} \\
		AP\_Omentum\_Kidney       & 0.0458          & 0.6731          & 0.0075          & \textbf{0.3087} & \textbf{0.8281} & \textbf{0.2801} \\
		AP\_Colon\_Lung           & 0.2111          & 0.6739          & 0.1180          & \textbf{0.2464} & \textbf{0.7060} & \textbf{0.1490} \\
		AP\_Colon\_Kidney         & \textbf{0.2975} & 0.6494          & 0.2280          & 0.2780          & \textbf{0.6838} & \textbf{0.3589} \\
		AP\_Endometrium\_Breast   & -0.0233         & 0.8447          & 0.0101          & \textbf{0.5397} & \textbf{0.9108} & \textbf{0.4453} \\
		AP\_Ovary\_Lung           & 0.0811          & 0.5567          & 0.0902          & \textbf{0.3552} & \textbf{0.7407} & \textbf{0.3229} \\
		AP\_Endometrium\_Prostate & -0.0058         & 0.5337          & 0.0007          & \textbf{0.6490} & \textbf{0.8245} & \textbf{0.6302} \\
		AP\_Endometrium\_Kidney   & -0.0059         & 0.8264          & 0.0032          & \textbf{0.4691} & \textbf{0.8781} & \textbf{0.3966} \\
		AP\_Ovary\_Kidney         & 0.0018          & 0.7122          & 0.0066          & \textbf{0.3655} & \textbf{0.7129} & \textbf{0.3112} \\
		AP\_Breast\_Ovary         & 0.0262          & 0.7291          & 0.0207          & \textbf{0.2927} & \textbf{0.7340} & \textbf{0.2261} \\
		AP\_Lung\_Uterus          & 0.0682          & 0.5333          & 0.0541          & \textbf{0.3330} & \textbf{0.6649} & \textbf{0.2587} \\
		AP\_Ovary\_Kidney         & 0.0018          & 0.7122          & 0.0066          & \textbf{0.3655} & \textbf{0.7129} & \textbf{0.3112} \\
		\midrule
		Average                   & 0.1559          & 0.6471          & 0.1790          & \textbf{0.3718} & \textbf{0.7113} & \textbf{0.3777} \\
		Median                    & 0.0729          & 0.6735          & 0.0722          & \textbf{0.3441} & \textbf{0.7129} & \textbf{0.3171} \\
		\bottomrule
	\end{tabular}	
	\label{tab:results2}
\end{table*}

A striking observation is that GTSA-PCA consistently outperforms SPCA across the majority of datasets and evaluation metrics, despite relying on only a small fraction of labeled data for hyperparameter tuning. This trend is particularly pronounced in high-dimensional and biologically motivated datasets (e.g., the \textit{AP-*} collection), where GTSA-PCA yields substantial gains in all three indices, often transforming near-random clustering performance (ARI $\approx 0$) into meaningful structure recovery. These results strongly suggest that, in such regimes, capturing the intrinsic geometry of the data manifold is more critical than enforcing global label-driven projections.

From a quantitative perspective, the improvements are significant and consistent: the average ARI increases from $0.1559$ (SPCA) to $0.3718$ (GTSA-PCA), while the V-measure more than doubles (from $0.1790$ to $0.3777$). The median values follow the same trend, indicating that the gains are not driven by a few favorable cases but are broadly distributed across datasets. Notably, GTSA-PCA exhibits remarkable robustness in challenging scenarios characterized by high dimensionality, small sample sizes, and complex nonlinear structure, where SPCA tends to struggle due to its reliance on global linear projections even when supervision is available.

There are, however, a few datasets (e.g., \textit{ionosphere} and \textit{iris}) where SPCA achieves competitive or slightly superior performance in specific metrics, particularly the Fowlkes-Mallows index. These cases typically correspond to datasets with relatively low intrinsic curvature or class structures that are well aligned with global variance directions, where the additional geometric modeling introduced by GTSA-PCA offers limited advantage. Nevertheless, even in these scenarios, GTSA-PCA remains highly competitive and often achieves the best performance in at least one complementary metric, such as V-measure.

Overall, these results highlight a key insight: a modest amount of supervision, when combined with a principled geometric framework, can outperform fully supervised linear methods. GTSA-PCA leverages label information indirectly, guiding model selection rather than dictating the embedding, while preserving the intrinsic manifold structure through curvature-aware local modeling and geodesic alignment. This balance between geometry and supervision appears to be particularly effective in complex, high-dimensional settings, reinforcing the central premise of this work.

\subsection{GTSA-PCA vs. PCA, Kernel PCA and Supervised PCA with HDBSCAN}

In the second set of experiments, we evaluate the quality of the low-dimensional representations under a density-based clustering regime by employing HDBSCAN. In contrast to hierarchical clustering with Ward linkage, HDBSCAN does not assume convex or spherical cluster structures and is capable of identifying clusters with varying densities while explicitly modeling noise. This makes it particularly suitable for assessing whether the embeddings produced by each method preserve the intrinsic local organization of the data. As in the previous setting, the data are projected onto a two-dimensional space using PCA, Kernel PCA, Supervised PCA, and GTSA-PCA prior to clustering. This experimental configuration provides a complementary perspective, allowing us to analyze how well each method captures nonlinear structures and density variations, which are critical aspects in manifold-based representations. Table~\ref{tab:results3} reports the clustering performance obtained using HDBSCAN on the two-dimensional embeddings produced by Regular PCA, Kernel PCA, and the proposed GTSA-PCA.

\begin{table*}
	\centering
	\caption{External indices obtained after HDBSCAN clustering in the two-dimensional features spaces generated by regular PCA, Kernel PCA and the proposed GTSA-PCA.}
	\begin{tabular}{cccccccccc}
		\toprule
		& \multicolumn{3}{c}{\textbf{Regular PCA}}     & \multicolumn{3}{c}{\textbf{Kernel PCA (RBF)}}       & \multicolumn{3}{c}{\textbf{GTSA-PCA}}               \\
		\midrule
		\textbf{Datasets}       & \textbf{ARI} & \textbf{FM}     & \textbf{VM} & \textbf{ARI}    & \textbf{FM}     & \textbf{VM}     & \textbf{ARI}    & \textbf{FM}     & \textbf{VM}     \\
		\midrule
		wine                    & 0.2729       & 0.5076          & 0.4205      & 0.5931          & 0.7175          & 0.6540          & \textbf{0.6744} & \textbf{0.7766} & \textbf{0.6773} \\
		breast\_cancer          & 0.4494       & 0.7297          & 0.3518      & 0.5027          & 0.7358          & 0.4161          & \textbf{0.5893} & \textbf{0.7919} & \textbf{0.4944} \\
		Engine1                 & 0.0838       & 0.4959          & 0.1799      & 0.1578          & 0.4436          & 0.2644          & \textbf{0.2658} & \textbf{0.5621} & \textbf{0.3159} \\
		user-knowledge          & -0.0114      & 0.3871          & 0.0817      & 0.0322          & 0.3220          & 0.1417          & \textbf{0.1968} & \textbf{0.4466} & \textbf{0.2646} \\
		mfeat-karhunen          & 0.0028       & 0.2682          & 0.0206      & 0.0160          & \textbf{0.3142} & 0.1325          & \textbf{0.1120} & 0.2852          & \textbf{0.2737} \\
		mfeat-factors           & 0.0139       & 0.2943          & 0.0868      & 0.0536          & 0.3100          & 0.2129          & \textbf{0.2002} & \textbf{0.3532} & \textbf{0.4029} \\
		mfeat-pixel             & 0.0054       & 0.2229          & 0.0327      & 0.0269          & 0.3052          & 0.1546          & \textbf{0.0589} & \textbf{0.3115} & \textbf{0.2116} \\
		cardiotocography        & 0.0501       & 0.4057          & 0.1182      & 0.0511          & 0.3970          & 0.1977          & \textbf{0.6807} & \textbf{0.7298} & \textbf{0.7464} \\
		pendigits (25\%)        & 0.0637       & 0.3119          & 0.2603      & 0.0870          & 0.3493          & 0.2784          & \textbf{0.3180} & \textbf{0.4359} & \textbf{0.5270} \\
		Fashion-MNIST           & 0.0013       & 0.2761          & 0.0203      & 0.0264          & \textbf{0.3190} & 0.1733          & \textbf{0.0956} & 0.2630          & \textbf{0.2604} \\
		UMIST\_Faces\_Cropped   & 0.0097       & 0.1721          & 0.0552      & \textbf{0.0485} & \textbf{0.2070} & 0.1690          & 0.0317          & 0.2069          & \textbf{0.1734} \\
		page-blocks (50\%)      & 0.1400       & 0.8469          & 0.0498      & 0.2114          & 0.6911          & \textbf{0.1726} & \textbf{0.3234} & \textbf{0.9017} & 0.1389          \\
		heart-c                 & 0.0852       & 0.5133          & 0.1354      & 0.1191          & 0.5328          & 0.1065          & \textbf{0.1782} & \textbf{0.5590} & \textbf{0.1857} \\
		heart-h                 & 0.1466       & 0.5244          & 0.1287      & -0.0028         & 0.4295          & 0.0515          & \textbf{0.3065} & \textbf{0.6620} & \textbf{0.2074} \\
		AP\_Omentum\_Kidney     & 0.0000       & \textbf{0.8006} & 0.0000      & 0.0896          & 0.5417          & 0.0512          & \textbf{0.1191} & 0.5992          & \textbf{0.1235} \\
		AP\_Colon\_Kidney       & -0.0023      & 0.5475          & 0.0007      & 0.0104          & 0.5002          & 0.0542          & \textbf{0.2662} & \textbf{0.5831} & \textbf{0.2841} \\
		AP\_Ovary\_Lung         & -0.0116      & 0.5286          & 0.0136      & -0.0069         & 0.5044          & 0.0032          & \textbf{0.1957} & \textbf{0.6202} & \textbf{0.1609} \\
		AP\_Breast\_Ovary       & 0.0127       & 0.5516          & 0.0022      & 0.0120          & 0.4561          & 0.0237          & \textbf{0.1132} & \textbf{0.5978} & \textbf{0.0758} \\
		AP\_Colon\_Lung         & -0.0283      & 0.5964          & 0.0124      & 0.0864          & 0.5384          & 0.0298          & \textbf{0.1762} & \textbf{0.6254} & \textbf{0.0873} \\
		AP\_Endometrium\_Breast & 0.0000       & 0.8617          & 0.0000      & 0.0820          & 0.6451          & 0.0462          & \textbf{0.6342} & \textbf{0.8951} & \textbf{0.4476} \\
		\midrule
		Average                 & 0.0642       & 0.4921          & 0.0985      & 0.1098          & 0.4630          & 0.1667          & \textbf{0.2768} & \textbf{0.5603} & \textbf{0.3029} \\
		Median                  & 0.0112       & 0.5104          & 0.0525      & 0.0523          & 0.4499          & 0.1481          & \textbf{0.1985} & \textbf{0.5904} & \textbf{0.2625} \\
		\bottomrule
	\end{tabular}
	\label{tab:results3}
\end{table*}

The results provide compelling evidence that GTSA-PCA yields substantially more informative representations under a density-based clustering paradigm. In particular, the proposed method achieves the highest average performance across all three evaluation metrics, with a pronounced improvement in Adjusted Rand Index (ARI) and V-measure (VM), indicating a significantly better recovery of the underlying class structure and a more faithful preservation of both cluster homogeneity and completeness.

A key observation is that GTSA-PCA consistently outperforms both PCA and Kernel PCA across the majority of datasets, especially those characterized by nonlinear structure and heterogeneous densities, such as \textit{cardiotocography}, \textit{pendigits}, \textit{Fashion-MNIST}, and several high-dimensional biological datasets from the \textit{AP-*} family. In these cases, the gains are often dramatic, transforming near-random clustering performance (ARI close to zero or negative) into meaningful partitions with clear alignment to ground-truth labels. This behavior highlights the importance of explicitly modeling both curvature and geodesic relationships: while PCA fails due to its global linearity, and Kernel PCA only partially captures nonlinearity through implicit mappings, GTSA-PCA constructs embeddings that respect the intrinsic geometry of the data manifold, which is crucial for density-based methods such as HDBSCAN.

Another important aspect is the robustness of GTSA-PCA in challenging scenarios involving high dimensionality and small sample sizes, as observed in the \textit{AP-*} datasets. In these settings, traditional methods often produce embeddings that collapse local structure or mix distinct clusters, leading to poor performance under HDBSCAN. In contrast, the curvature-aware local modeling and geodesic alignment of GTSA-PCA preserve meaningful neighborhood relationships, enabling HDBSCAN to effectively identify clusters and separate noise.

Although Kernel PCA occasionally achieves competitive results, particularly in terms of the Fowlkes-Mallows (FM) index on certain datasets, it generally falls short in ARI and VM, suggesting that the clusters it produces are less consistent with the true class assignments. Similarly, PCA exhibits limited effectiveness across most datasets, reinforcing its inadequacy for capturing nonlinear and density-varying structures.

The aggregate statistics further confirm these trends: GTSA-PCA more than doubles the average ARI compared to PCA and significantly improves upon Kernel PCA, while also achieving the highest median values across all metrics. These consistent gains across both average and median statistics indicate that the improvements are stable and not driven by isolated cases. Overall, the results demonstrate that embeddings produced by GTSA-PCA are particularly well-suited for density-based clustering, as they preserve the local and global geometric properties necessary for accurately capturing complex data distributions.


The results presented in Table~\ref{tab:results4} provide a detailed comparison between Supervised PCA and the proposed GTSA-PCA when combined with HDBSCAN clustering. Overall, the proposed method consistently outperforms Supervised PCA across the majority of datasets and evaluation metrics. This superiority is reflected both in the average scores, ARI (0.3041 vs. 0.1972), FM (0.5794 vs. 0.5200), and V-measure (0.3465 vs. 0.2724), and in the median values, indicating that the observed gains are robust and not driven by isolated cases.

\begin{table*}
	\centering
	\caption{External indices obtained after HDBSCAN clustering in the two-dimensional features spaces generated by Supervised PCA and the proposed GTSA-PCA.}
	\begin{tabular}{ccccccc}
		\toprule
		& \multicolumn{3}{c}{\textbf{Supervised PCA}}      & \multicolumn{3}{c}{\textbf{GTSA-PCA}}               \\
		\midrule
		\textbf{Datasets}   & \textbf{ARI} & \textbf{FM}     & \textbf{VM}     & \textbf{ARI}    & \textbf{FM}     & \textbf{VM}     \\
		\midrule
		iris                 & 0.5908          & 0.7742          & 0.7274          & \textbf{0.5949} & \textbf{0.7829} & \textbf{0.7458} \\
		breast\_cancer       & 0.4803          & 0.7346          & 0.3889          & \textbf{0.5893} & \textbf{0.7918} & \textbf{0.4943} \\
		Engine1              & 0.0593          & 0.4786          & 0.2146          & \textbf{0.2658} & \textbf{0.5621} & \textbf{0.3159} \\
		cardiotocography     & 0.5019          & 0.6224          & 0.5716          & \textbf{0.6807} & \textbf{0.7298} & \textbf{0.7464} \\
		mfeat-zernike        & \textbf{0.0338} & 0.2850          & \textbf{0.1458} & 0.0312          & \textbf{0.2911} & 0.1429          \\
		thyroid-new          & 0.0000          & 0.7371          & 0.0000          & \textbf{0.5075} & \textbf{0.7662} & \textbf{0.3463} \\
		seeds                & 0.3312          & 0.5351          & 0.4633          & \textbf{0.4765} & \textbf{0.6319} & \textbf{0.5248} \\
		xd6                  & 0.0203          & 0.1873          & \textbf{0.1233} & \textbf{0.0523} & \textbf{0.5246} & 0.0340          \\
		cars1                & 0.0218          & 0.3738          & 0.1510          & \textbf{0.0529} & \textbf{0.4439} & \textbf{0.2342} \\
		threeOf9             & 0.0042          & 0.4882          & 0.0307          & \textbf{0.0144} & \textbf{0.5374} & \textbf{0.0194} \\
		corral               & 0.0805          & 0.4651          & 0.1359          & \textbf{0.3115} & \textbf{0.6417} & \textbf{0.2278} \\
		wine                 & 0.5846          & 0.7116          & 0.6425          & \textbf{0.6744} & \textbf{0.7766} & \textbf{0.6773} \\
		Breast               & -0.0446         & 0.4573          & 0.1617          & \textbf{0.1754} & \textbf{0.4664} & \textbf{0.3037} \\
		ionosphere           & \textbf{0.1116} & \textbf{0.4819} & 0.1877          & 0.1103          & 0.4286          & \textbf{0.2274} \\
		heart-h              & 0.2860          & 0.6248          & 0.1885          & \textbf{0.3065} & \textbf{0.6620} & \textbf{0.2074} \\
		Fashion-MNIST (5\%)  & 0.0361          & \textbf{0.3309} & 0.2053          & \textbf{0.0956} & 0.2630          & \textbf{0.2604} \\
		breast\_cancer       & 0.4803          & 0.7346          & 0.3889          & \textbf{0.5893} & \textbf{0.7919} & \textbf{0.4944} \\
		user-knowledge       & 0.0848          & \textbf{0.4735} & 0.1646          & \textbf{0.1968} & 0.4466          & \textbf{0.2646} \\
		pendigits (25\%)     & 0.2978          & 0.4065          & 0.5146          & \textbf{0.3180} & \textbf{0.4359} & \textbf{0.5270} \\
		arsenic-male-bladder & -0.0172         & 0.4967          & 0.0423          & \textbf{0.0382} & \textbf{0.6149} & \textbf{0.1352} \\
		\midrule
		Average              & 0.1972          & 0.5200          & 0.2724          & \textbf{0.3041} & \textbf{0.5794} & \textbf{0.3465} \\
		Median               & 0.0827          & 0.4850          & 0.1881          & \textbf{0.2862} & \textbf{0.5885} & \textbf{0.2841} \\
		\bottomrule
	\end{tabular}
	\label{tab:results4}
\end{table*}

A closer inspection reveals that GTSA-PCA achieves substantial improvements in datasets characterized by nonlinear structure or complex class distributions. Notable examples include \textit{Engine1}, \textit{cardiotocography}, \textit{thyroid-new}, and \textit{corral}, where the gains in ARI and V-measure are particularly significant. These results suggest that the proposed method is more effective at preserving the intrinsic geometry of the data, leading to embeddings that better reflect the true cluster structure and are therefore more amenable to density-based clustering.

Interestingly, even in scenarios where Supervised PCA has access to full label information, GTSA-PCA is often able to surpass its performance. This highlights a key limitation of Supervised PCA: although it leverages supervision, it remains fundamentally constrained by a global linear projection model. As a result, it may fail to capture nonlinear relationships and geometric distortions present in the data. In contrast, GTSA-PCA incorporates local geometric information through tangent space estimation and global consistency via spectral alignment, enabling it to better model curved manifolds.

There are a few cases where Supervised PCA remains competitive or slightly superior in specific metrics, such as in \textit{mfeat-zernike} and \textit{ionosphere}. These datasets likely exhibit structures that are closer to linear separability or where class information is well aligned with variance directions, favoring the assumptions underlying Supervised PCA. Nevertheless, even in these cases, GTSA-PCA remains competitive and often achieves better performance in at least one of the evaluated metrics.

Another important aspect is the compatibility between GTSA-PCA and HDBSCAN. Since HDBSCAN relies on local density estimation and is sensitive to the topology of the embedding space, the geometry-aware representations produced by GTSA-PCA provide a more suitable input for the clustering algorithm. This is reflected in the consistent improvements observed across datasets, particularly in terms of ARI, which measures agreement with the ground-truth partition.

In summary, the results demonstrate that GTSA-PCA not only provides a robust alternative to Supervised PCA but can also outperform it even in the presence of full supervision. This reinforces the central premise of this work: explicitly modeling the geometric structure of the data manifold is a powerful strategy for dimensionality reduction, often more effective than relying solely on label information within a global linear framework.

\subsection{GTSA-PCA-W vs. UMAP with agglomerative and HDBSCAN clustering}

The third set of experiments aims to evaluate the effectiveness of the proposed framework under its Wasserstein-based variant, in which curvature-dependent weights are replaced by Wasserstein distances, and to compare it against UMAP, a state-of-the-art manifold learning method widely recognized for its ability to preserve both local and global structures. This comparison is conducted under two clustering regimes, agglomerative clustering with Ward linkage and HDBSCAN, in order to assess the robustness of the learned representations across different notions of cluster structure. The motivation for this experiment is twofold: first, to investigate whether replacing curvature with an optimal transport-based metric yields more stable and scalable behavior in high-dimensional settings; and second, to benchmark the proposed method against a strong nonlinear baseline that is explicitly designed for low-dimensional embedding. By evaluating clustering quality through multiple external indices, we aim to provide a comprehensive assessment of how well each method captures the intrinsic geometry and class structure of the data. Table \ref{tab:results5} shows the obtained results.

\begin{table*}
	\centering
	\caption{External indices obtained after agglomerative clustering in the two-dimensional features spaces generated by UMAP and the proposed GTSA-PCA with Wasserstein distance.}
	\begin{tabular}{ccccccc}
		\toprule
		& \multicolumn{3}{c}{\textbf{UMAP}}                   & \multicolumn{3}{c}{\textbf{GTSA-PCA-W}}             \\
		\midrule
		\textbf{Datasets}    & \textbf{ARI}    & \textbf{FM}     & \textbf{VM}     & \textbf{ARI}    & \textbf{FM}     & \textbf{VM}     \\
		\midrule
		breast\_cancer       & 0.3758          & 0.6988          & 0.3405          & \textbf{0.5970} & \textbf{0.8227} & \textbf{0.4886} \\
		AP\_Lung\_Kidney     & 0.1106          & 0.5951          & 0.2176          & \textbf{0.3194} & \textbf{0.7576} & \textbf{0.2565} \\
		AP\_Colon\_Lung      & 0.2760          & 0.6594          & 0.2491          & \textbf{0.3641} & \textbf{0.7503} & \textbf{0.2536} \\
		AP\_Ovary\_Kidney    & 0.4294          & 0.7204          & 0.4260          & \textbf{0.4440} & \textbf{0.7270} & \textbf{0.4356} \\
		OVA\_Kidney          & 0.2025          & 0.8577          & 0.1660          & \textbf{0.7455} & \textbf{0.9271} & \textbf{0.5737} \\
		Breast               & 0.2029          & 0.6251          & 0.2483          & \textbf{0.6929} & \textbf{0.8666} & \textbf{0.5704} \\
		Dexter               & \textbf{0.1327} & 0.5700          & 0.1021          & 0.0679          & \textbf{0.6374} & \textbf{0.1463} \\
		ionosphere           & 0.0038          & 0.5431          & 0.0392          & \textbf{0.1589} & \textbf{0.5921} & \textbf{0.1301} \\
		Engine1              & -0.0252         & 0.5884          & 0.1575          & \textbf{0.0953} & \textbf{0.5954} & \textbf{0.2493} \\
		diabetes             & 0.0082          & \textbf{0.7064} & 0.0012          & \textbf{0.1008} & 0.5731          & \textbf{0.0751} \\
		glass                & 0.1357          & 0.3350          & 0.2443          & \textbf{0.1804} & \textbf{0.3905} & \textbf{0.3157} \\
		ecoli                & 0.3652          & 0.5111          & \textbf{0.6010} & \textbf{0.3930} & \textbf{0.5368} & 0.5612          \\
		heart-h              & 0.2783          & 0.6653          & 0.1901          & \textbf{0.3168} & \textbf{0.6799} & \textbf{0.2231} \\
		heart-c              & 0.2208          & 0.6107          & 0.1684          & \textbf{0.4045} & \textbf{0.7058} & \textbf{0.3300} \\
		accute-inflammations & 0.0225          & 0.5419          & 0.0583          & \textbf{0.4067} & \textbf{0.7499} & \textbf{0.4285} \\
		liver-disorders      & 0.0066          & 0.1133          & 0.1310          & \textbf{0.0191} & \textbf{0.1475} & \textbf{0.1410} \\
		colic                & 0.0131          & 0.5602          & 0.0728          & \textbf{0.0819} & \textbf{0.5626} & \textbf{0.1053} \\
		MegaWatt             & 0.0357          & 0.6460          & 0.0522          & \textbf{0.1640} & \textbf{0.7588} & \textbf{0.0693} \\
		threeOf9             & 0.0572          & 0.5281          & 0.0432          & \textbf{0.3353} & \textbf{0.6673} & \textbf{0.2595} \\
		corral               & 0.2766          & 0.6399          & 0.2107          & \textbf{0.3109} & \textbf{0.6615} & \textbf{0.2356} \\
		\midrule
		Average              & 0.1564          & 0.5858          & 0.1860          & \textbf{0.3099} & \textbf{0.6555} & \textbf{0.2924} \\
		Median               & 0.1342          & 0.6029          & 0.1672          & \textbf{0.3181} & \textbf{0.6736} & \textbf{0.2551} \\
		\bottomrule
	\end{tabular}
	\label{tab:results5}
\end{table*}

The results provide a comprehensive comparison between the proposed Wasserstein-based variant of GTSA-PCA and UMAP under agglomerative clustering. Overall, the proposed method demonstrates competitive, and in several cases superior, performance across the evaluated datasets, particularly in terms of ARI and V-measure, indicating a more consistent preservation of the underlying cluster structure after dimensionality reduction.

A key observation concerns the behavior of UMAP in small sample size regimes. While UMAP is well known for its strong performance on large-scale datasets due to its ability to capture both local and global manifold structure, it exhibits a notable limitation when the number of samples is limited. In such scenarios, the construction of a reliable fuzzy simplicial complex becomes challenging, leading to unstable neighborhood graphs and, consequently, embeddings that may not faithfully reflect the intrinsic geometry of the data. This effect is evident in several datasets with relatively few instances, where UMAP underperforms compared to the proposed method, particularly in ARI scores, suggesting poorer alignment with ground-truth cluster assignments.

In contrast, the Wasserstein-based GTSA-PCA maintains a more stable performance across varying sample sizes. By leveraging optimal transport distances to define neighborhood relationships, the method provides a more robust notion of similarity that is less sensitive to sampling sparsity. This results in embeddings that better preserve both local consistency and global structure, which is further reflected in improved clustering outcomes under Ward linkage. The advantage is especially pronounced in datasets where the geometric structure is complex but the number of samples is insufficient for reliable density estimation, a scenario in which UMAP tends to struggle.

The results reported in Table~\ref{tab:results6} provide a clear and consistent comparison between UMAP and the proposed Wasserstein-based GTSA-PCA (GTSA-PCA-W) when coupled with HDBSCAN clustering. Overall, the proposed method substantially outperforms UMAP across all evaluation metrics, with significant gains in ARI (0.3868 vs. 0.1051), FM (0.6846 vs. 0.5113), and V-measure (0.3571 vs. 0.1917) in terms of average performance. The median values further reinforce this trend, indicating that the improvements are not driven by isolated cases but rather reflect a systematic advantage across datasets.

\begin{table*}
	\centering
	\caption{External indices obtained after HDBSCAN clustering in the two-dimensional features spaces generated by UMAP and the proposed GTSA-PCA with Wasserstein distance.}
	\begin{tabular}{ccccccc}
		\toprule
		& \multicolumn{3}{c}{\textbf{UMAP}}            & \multicolumn{3}{c}{\textbf{GTSA-PCA-W}}             \\
		\midrule
		\textbf{Datasets}   & \textbf{ARI} & \textbf{FM}     & \textbf{VM} & \textbf{ARI}    & \textbf{FM}     & \textbf{VM}     \\
		\midrule
		AP\_Lung\_Kidney    & -0.0723      & 0.6350          & 0.0746      & \textbf{0.4912} & \textbf{0.7531} & \textbf{0.3906} \\
		AP\_Colon\_Lung     & 0.1311       & 0.5227          & 0.1638      & \textbf{0.6476} & \textbf{0.8316} & \textbf{0.4854} \\
		AP\_Ovary\_Kidney   & -0.0187      & 0.6269          & 0.0659      & \textbf{0.4891} & \textbf{0.7153} & \textbf{0.4333} \\
		AP\_Prostate\_Ovary & 0.3051       & 0.6577          & 0.3693      & \textbf{0.7166} & \textbf{0.8862} & \textbf{0.6284} \\
		OVA\_Kidney         & 0.0613       & 0.3379          & 0.2366      & \textbf{0.6389} & \textbf{0.8948} & \textbf{0.4453} \\
		Breast              & 0.1566       & 0.4519          & 0.2892      & \textbf{0.7138} & \textbf{0.8642} & \textbf{0.5856} \\
		diabetes            & 0.0160       & 0.5565          & 0.0044      & \textbf{0.0297} & \textbf{0.6728} & \textbf{0.0067} \\
		heart-h             & 0.1496       & 0.5280          & 0.2093      & \textbf{0.3066} & \textbf{0.6672} & \textbf{0.2021} \\
		AP\_Uterus\_Kidney  & -0.0458      & 0.6414          & 0.0409      & \textbf{0.3928} & \textbf{0.6909} & \textbf{0.3978} \\
		AP\_Breast\_Lung    & 0.0350       & 0.5352          & 0.0780      & \textbf{0.1464} & \textbf{0.5984} & \textbf{0.0960} \\
		iris                & 0.4589       & 0.6813          & 0.6586      & \textbf{0.5949} & \textbf{0.7829} & \textbf{0.7458} \\
		ionosphere          & 0.1655       & 0.5025          & 0.2278      & \textbf{0.2481} & \textbf{0.5748} & \textbf{0.2496} \\
		Engine1             & 0.0865       & \textbf{0.5238} & 0.2262      & \textbf{0.1025} & 0.4686          & \textbf{0.2304} \\
		tic-tac-toe         & 0.1021       & 0.4286          & 0.1581      & \textbf{0.4741} & \textbf{0.7063} & \textbf{0.5565} \\
		threeOf9            & 0.0141       & 0.2893          & 0.0608      & \textbf{0.0712} & \textbf{0.3140} & \textbf{0.1620} \\
		corral              & 0.2766       & 0.6399          & 0.2107      & \textbf{0.3483} & \textbf{0.6476} & \textbf{0.2557} \\
		page-blocks (50\%)  & 0.0266       & 0.3004          & 0.1488      & \textbf{0.4251} & \textbf{0.9078} & \textbf{0.2160} \\
		Fashion-MNIST (5\%) & 0.0436       & 0.3446          & 0.2274      & \textbf{0.1256} & \textbf{0.3466} & \textbf{0.3398} \\
		\midrule
		Average             & 0.1051       & 0.5113          & 0.1917      & \textbf{0.3868} & \textbf{0.6846} & \textbf{0.3571} \\
		Median              & 0.0739       & 0.5259          & 0.1866      & \textbf{0.4090} & \textbf{0.6986} & \textbf{0.3652} \\
		\bottomrule
	\end{tabular}
	\label{tab:results6}
\end{table*}

A closer inspection reveals that GTSA-PCA-W achieves superior performance in nearly all datasets, often with large margins. This is particularly evident in several high-dimensional biomedical datasets (e.g., \textit{AP\_Lung\_Kidney}, \textit{AP\_Colon\_Lung}, and \textit{OVA\_Kidney}), where UMAP struggles to produce embeddings that are well-aligned with the underlying class structure, as reflected by low or even negative ARI scores. In contrast, the proposed method yields substantially higher ARI, FM, and V-measure values, suggesting that the Wasserstein-based neighborhood modeling combined with spectral tangent space aggregation is more effective at preserving meaningful geometric relationships in such settings.

Furthermore, the combination of Wasserstein distances with the spectral aggregation framework appears to yield embeddings that are particularly well-suited for density-based clustering. This is evidenced by the consistent gains observed under HDBSCAN, where the proposed method more effectively separates clusters with varying densities and irregular shapes. In summary, while UMAP remains a strong baseline in large-scale settings, the proposed approach offers a more robust and reliable alternative in small to medium-sized datasets, where geometric consistency and stability are critical.

Finally, it is worth noting that even in datasets where UMAP remains competitive in specific metrics (e.g., FM in \textit{Engine1}), the proposed method still achieves comparable or superior performance in ARI and V-measure, indicating better overall agreement with the ground-truth partition. In summary, the results strongly suggest that GTSA-PCA-W provides a more reliable and geometry-aware alternative to UMAP, particularly in challenging scenarios involving high-dimensional, small-sample, or nonlinearly structured data.

\subsection{Qualitative results}

In addition to the quantitative evaluation, we present a qualitative analysis of the low-dimensional embeddings produced by the different methods considered in this work. Specifically, this subsection illustrates a series of two-dimensional scatter plots obtained after dimensionality reduction, allowing for a direct visual comparison of how each method organizes the data in the embedding space. The goal is to assess, from a geometric and structural perspective, the extent to which class separability, cluster compactness, and manifold continuity are preserved. Such qualitative inspection complements the numerical results by revealing properties that are not always fully captured by external clustering indices, such as the presence of distortions, overlapping regions, or artificial fragmentation of clusters. This combined analysis provides a more comprehensive understanding of the strengths and limitations of each approach.

\begin{figure}[H]
	\centering
	\includegraphics[scale=0.28]{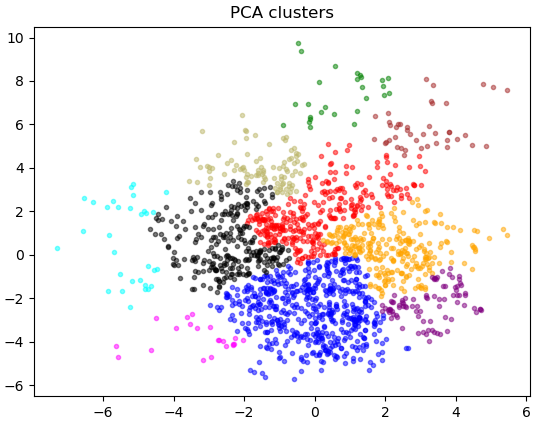}
	\includegraphics[scale=0.28]{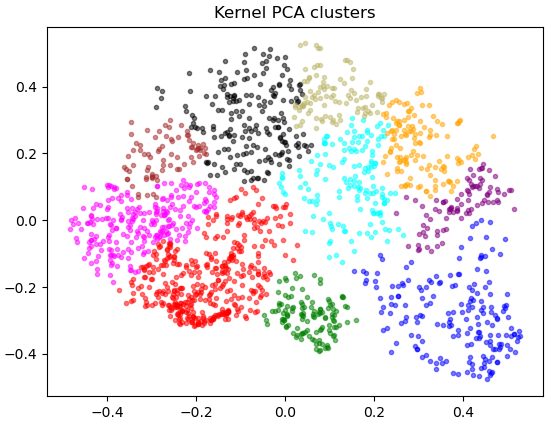}
	\includegraphics[scale=0.28]{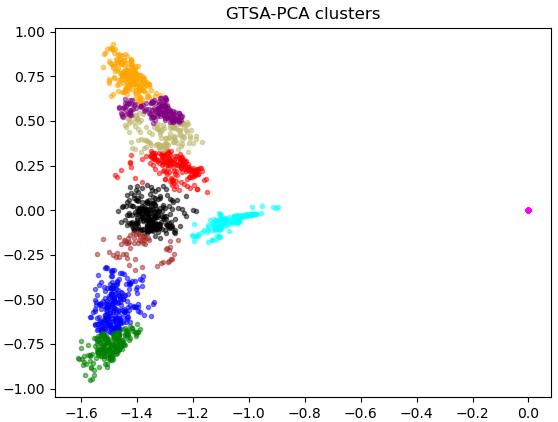}
	\caption{Scatterplots obtained after agglomerative clustering is applied to the reduced data in cardiotocography dataset. From left to right: regular PCA, Kernel PCA (rbf) and the proposed GTSA-PCA.}
	\label{fig:plot1}
\end{figure}

Figure \ref{fig:plot1} shows a comparison between the clusters obtained by agglomerative clustering after regular PCA, Kernel PCA (rbf) and the proposed GTSA-PCA applied to the cardiotocography dataset. Classical PCA yields a projection with substantial class overlap and poorly defined cluster boundaries, reflecting its inability to capture nonlinear structures. Kernel PCA partially alleviates this limitation by introducing nonlinearity through the RBF kernel, leading to more compact clusters; however, a noticeable degree of overlap remains, particularly in central regions, indicating that the intrinsic geometry is not fully preserved. In contrast, GTSA-PCA produces a significantly more organized embedding, with well-separated clusters, reduced intra-class dispersion, and enhanced inter-class margins. This behavior suggests that explicitly incorporating geometric information, via geodesic alignment and aggregation of local tangent spaces, enables a more faithful approximation of the underlying manifold, resulting in embeddings that are better suited for clustering tasks.

\begin{figure}[H]
	\centering
	\includegraphics[scale=0.28]{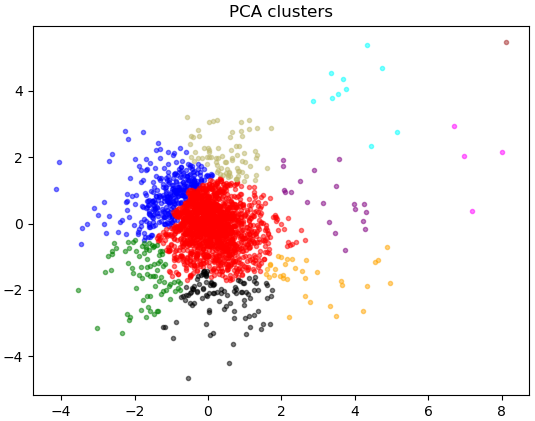}
	\includegraphics[scale=0.28]{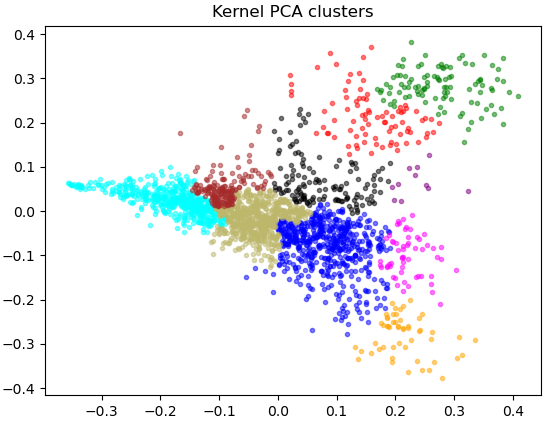}
	\includegraphics[scale=0.28]{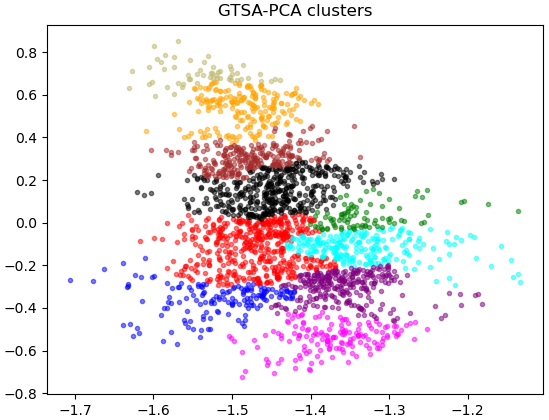}
	\caption{Scatterplots obtained after agglomerative clustering is applied to the reduced data in the optdigits dataset. From left to right: regular PCA, Kernel PCA (rbf) and the proposed GTSA-PCA.}
	\label{fig:plot2}
\end{figure}

Figure \ref{fig:plot2} shows a comparison between the clusters obtained by agglomerative clustering after regular PCA, Kernel PCA (rbf) and the proposed GTSA-PCA applied to the optdigits dataset. The scatter plots for the \textit{optdigits} dataset highlight a marked contrast between the embeddings produced by PCA and GTSA-PCA under agglomerative clustering. The PCA projection exhibits a highly entangled structure, where most classes collapse into a dense central region with significant overlap and poorly distinguishable boundaries, indicating a loss of discriminative information in the linear projection. In contrast, GTSA-PCA yields a substantially more organized representation, where clusters are more spatially separated and exhibit clearer geometric structure, even in the presence of moderate overlap. This improved arrangement suggests that the proposed method is more effective at unfolding the underlying manifold and preserving class-specific variations, leading to embeddings that are better aligned with the intrinsic data geometry and more suitable for clustering tasks.

\begin{figure}[H]
	\centering
	\includegraphics[scale=0.33]{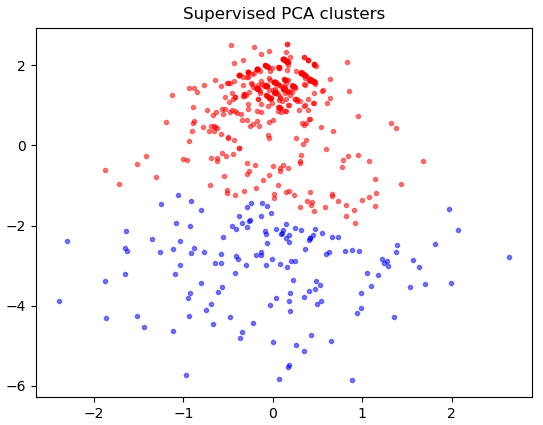}
	\includegraphics[scale=0.33]{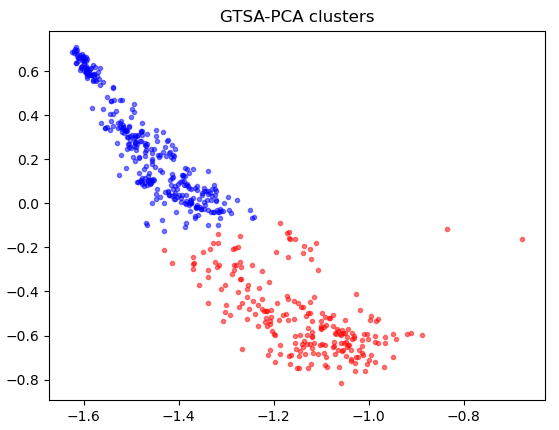}
	\caption{Scatterplots obtained after agglomerative clustering is applied to the reduced data in the Breast dataset. From left to right: Supervised PCA and the proposed GTSA-PCA.}
	\label{fig:plot3}
\end{figure}

Figure~\ref{fig:plot3} presents a qualitative comparison of the cluster structures obtained via agglomerative clustering applied to the embeddings produced by Supervised PCA and the proposed GTSA-PCA on the Breast dataset. Notably, Supervised PCA, despite having access to full label information during training, yields embeddings with considerably more dispersed and less cohesive cluster formations than those recovered by GTSA-PCA. This observation suggests that label supervision alone is insufficient to induce geometrically faithful representations, and that explicitly accounting for intrinsic geometric properties of the data manifold, such as local tangent space variation and curvature, is critical for learning embeddings that reflect meaningful 
structure. These results position GTSA-PCA as a principled alternative that achieves 
superior cluster separability not by relying on stronger supervisory signals, but by 
grounding the dimensionality reduction in the underlying geometry of the data. 

\begin{figure}[H]
	\centering
	\includegraphics[scale=0.28]{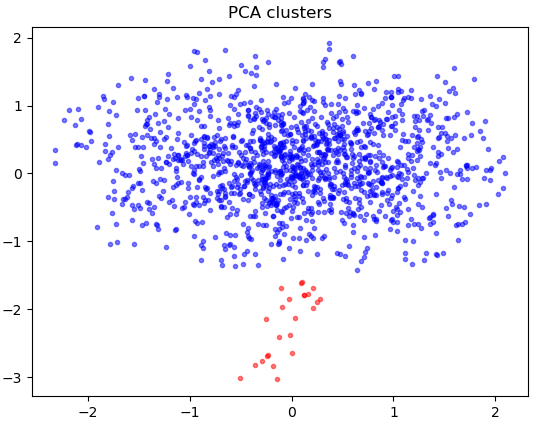}
	\includegraphics[scale=0.28]{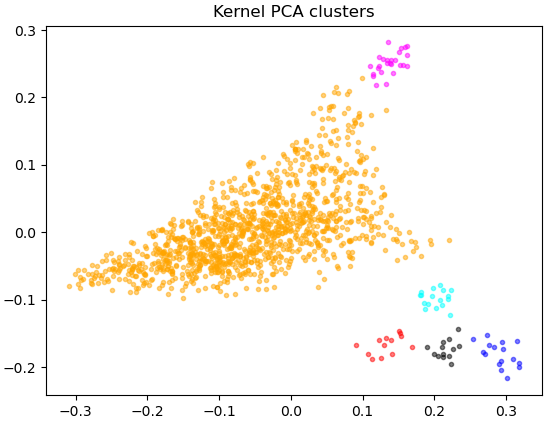}
	\includegraphics[scale=0.28]{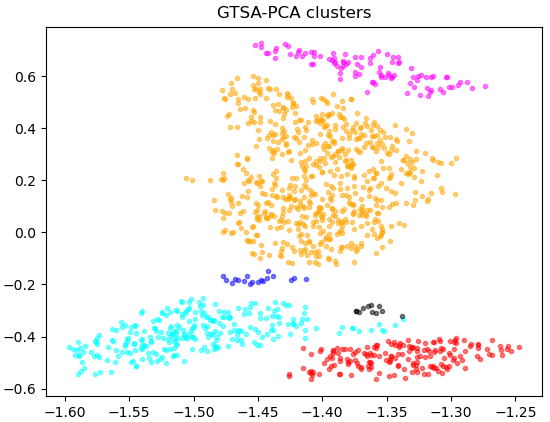}
	\caption{Scatterplots obtained after HDBSCAN clustering is applied to the reduced data in the mfeat-factors dataset. From left to right: regular PCA, Kernel PCA (rbf) and the proposed GTSA-PCA.}
	\label{fig:plot5}
\end{figure}

Figure \ref{fig:plot5} shows a comparison between the clusters obtained by HDBSCAN clustering after regular PCA, Kernel PCA (rbf) and the proposed GTSA-PCA applied to the mfeat-factors dataset. Standard PCA yields a largely undifferentiated embedding, collapsing most observations into a single dominant blue cluster while isolating only a small, loosely defined red group in the lower region of the projection, reflecting its inability to capture nonlinear geometric structure. Kernel PCA improves upon this by revealing additional minority clusters (magenta, cyan, gray, and blue) near the periphery of the embedding; however, the dominant orange cluster remains heavily  concentrated and poorly separated from the remaining groups, suggesting that the nonlinear kernel mapping alone is insufficient to disentangle the full manifold structure. In contrast, GTSA-PCA produces a markedly more structured embedding, in which six distinct clusters are clearly delineated across the projection space: a large central orange group, a magenta cluster in the upper region, a cyan group in the lower-left, a red cluster at the bottom-right, and two smaller compact groups (blue and gray) occupying intermediate positions. This separation demonstrates that the curvature-weighted local covariance operators and geodesic alignment introduced by GTSA-PCA effectively disentangle overlapping manifold structures that remain conflated under both PCA and Kernel PCA.

\begin{figure}[H]
	\centering
	\includegraphics[scale=0.33]{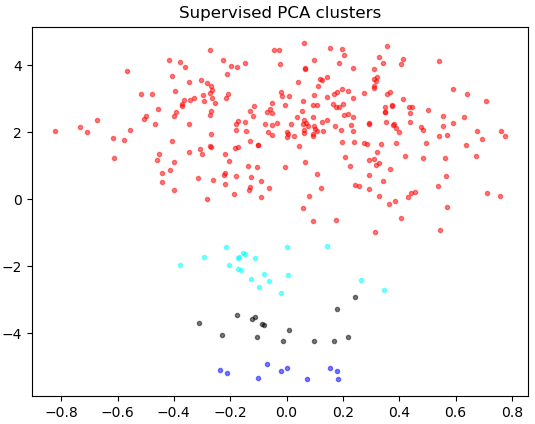}
	\includegraphics[scale=0.33]{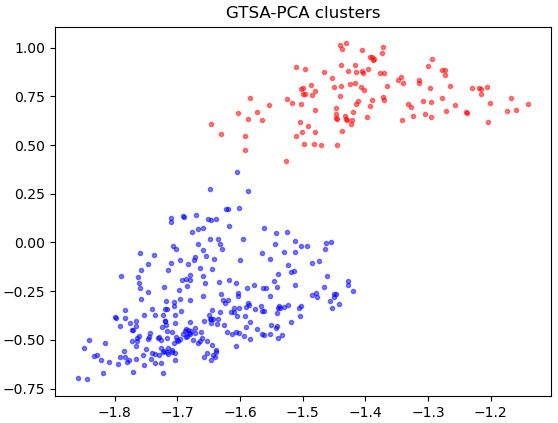}
	\caption{Scatterplots obtained after HDBSCAN clustering is applied to the reduced data in the breast\_cancer dataset. From left to right: Supervised PCA and the proposed GTSA-PCA.}
	\label{fig:plot6}
\end{figure}

Figure~\ref{fig:plot6} presents a qualitative comparison of the cluster structures recovered by HDBSCAN applied to the embeddings produced by Supervised PCA and the proposed GTSA-PCA on the \texttt{breast\_cancer} dataset. Although the ground truth of this dataset comprises exactly two classes, HDBSCAN operating on the Supervised PCA embedding identifies four distinct clusters, indicating that the resulting representation introduces artificial separations not present in the underlying class structure. In contrast, GTSA-PCA yields an embedding from which HDBSCAN recovers precisely two clusters, in exact agreement with the ground truth partition. This discrepancy highlights a fundamental limitation of supervision-driven dimensionality reduction: label information alone does not guarantee geometrically coherent embeddings. The ability of GTSA-PCA to recover the correct cluster structure without explicit label guidance underscores the role of intrinsic geometric properties (curvature) in producing representations that are faithful to the true data manifold.

\begin{figure}[H]
	\centering
	\includegraphics[scale=0.33]{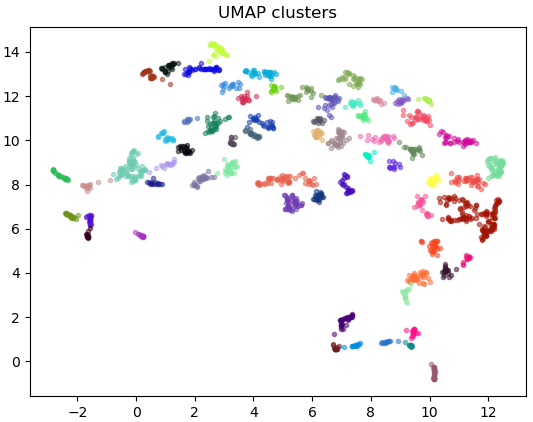}
	\includegraphics[scale=0.33]{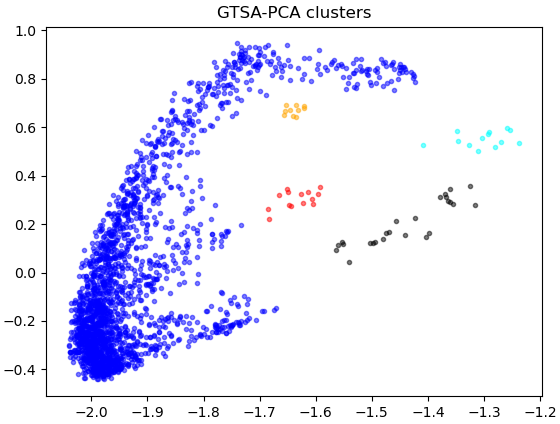}
	\caption{Scatterplots obtained after HDBSCAN clustering is applied to the reduced data in the page-blocks dataset. From left to right: UMAP and the proposed GTSA-PCA.}
	\label{fig:plot7}
\end{figure}

Finally, we consider the \texttt{page-blocks} dataset, which contains five ground-truth classes. When HDBSCAN is applied to the UMAP embedding, it identifies 64 distinct clusters: a severe overpartitioning that indicates the embedding fails to consolidate the underlying class structure into geometrically coherent regions. In contrast, HDBSCAN applied to the GTSA-PCA embedding recovers only four clusters, a result substantially closer to the true number of classes and suggestive of a far more faithful representation of the manifold geometry. Figure~\ref{fig:plot7} provides a qualitative illustration of this contrast, displaying the cluster structures recovered 
by HDBSCAN after Supervised PCA and after the proposed GTSA-PCA on the \texttt{page-blocks} dataset. The stark difference in cluster granularity between the two methods reinforces the broader argument that explicitly modeling curvature and local tangent space variation yields embeddings that are better aligned with the intrinsic structure of the data, even in the absence of label supervision.

For the interested reader, the Python source code to reproduce all the results shown in this paper can be found at: \url{https://github.com/alexandrelevada/GTSA-PCA}.

\section{Conclusions and Final Remarks}

In this work, we introduced GTSA-PCA, a geometric generalization of Principal Component Analysis that integrates curvature awareness and geodesic consistency into a unified spectral framework. The proposed method replaces the classical global covariance operator with a collection of curvature-regularized local covariance matrices, enabling the estimation of tangent spaces that better approximate the underlying manifold structure. These local representations are then globally aligned through a geodesic affinity operator that jointly accounts for intrinsic distances and subspace coherence. The resulting spectral decomposition yields low-dimensional embeddings that preserve both the local geometry and the global organization of the data. Extensive experiments on a large and diverse collection of real-world datasets demonstrated that GTSA-PCA consistently outperforms classical PCA, Kernel PCA, Supervised PCA, and UMAP, particularly in scenarios where nonlinear structure and curvature effects are pronounced.

An additional perspective of the proposed method is its interpretation as a novel approach to metric learning through dimensionality reduction. Unlike classical metric learning techniques that explicitly optimize pairwise or triplet-based objectives, GTSA-PCA implicitly induces a data-adaptive metric by constructing embeddings that respect both the local tangent geometry and the global geodesic structure of the data manifold. In this sense, the learned low-dimensional representation can be viewed as a transformation under which Euclidean distances approximate intrinsic geodesic distances while remaining consistent with curvature-aware local neighborhoods. This positions GTSA-PCA as a principled and unsupervised (or weakly supervised) alternative to traditional metric learning frameworks, bridging spectral methods, manifold learning, and geometric data analysis within a unified formulation.

Despite these advantages, the proposed method presents some limitations. First, the computational complexity is significantly higher than that of standard PCA, primarily due to the construction of local neighborhoods, curvature estimation, geodesic distance computation, and the spectral decomposition of a dense affinity matrix. This may limit scalability to very large datasets without additional approximations. Second, unlike classical PCA, GTSA-PCA does not naturally provide an explicit mapping for out-of-sample data, making its application in inductive settings less straightforward. Although extensions based on Nyström approximations or learned parametric mappings are possible, they are not inherent to the current formulation.

These limitations open several promising directions for future research. One avenue is the development of scalable approximations, such as sparse graph constructions, landmark-based methods, or randomized eigensolvers, to reduce computational cost. Another direction involves designing principled out-of-sample extensions, either through kernelization, neural approximators, or geometric interpolation schemes on the learned manifold. Additionally, improving the robustness of curvature estimation in high-dimensional or noisy regimes remains an important challenge, potentially benefiting from alternative geometric descriptors such as Wasserstein-based metrics or diffusion-based operators. Finally, extending the framework to incorporate stronger forms of supervision, temporal dynamics, or integration with deep representation learning architectures may further enhance its applicability and performance across a wider range of machine learning tasks.

\section*{Acknowledgments}
This work has been supported by CNPq (National Council for Scientific and Technological Development) through grant number 301432/2025-2. This study was also financed in part by the Coordenação de Aperfeiçoamento de Pessoal de N\'ivel Superior - Brasil (CAPES) - Finance Code 001.

\bibliography{main}

\end{document}

%% file: math_commands.tex
\usepackage{amsmath,amsfonts,bm}

\def\eqref#1{equation~\ref{#1}}

\def\1{\bm{1}}

\DeclareMathAlphabet{\mathsfit}{\encodingdefault}{\sfdefault}{m}{sl}
\SetMathAlphabet{\mathsfit}{bold}{\encodingdefault}{\sfdefault}{bx}{n}

